 \documentclass[preprint,review,12pt]{elsarticle}

\usepackage{amssymb}
\usepackage{subcaption}
\usepackage{graphicx}
\usepackage{verbatim}
\usepackage{mathtools}
\usepackage{hyperref}
\usepackage{mathrsfs}
\usepackage{enumitem}
\usepackage{color}
\usepackage{booktabs}
\usepackage{natbib}
\usepackage[algoruled,boxed,lined]{algorithm2e}
\usepackage{algorithmic}

\def\bf#1{\boldsymbol{#1}}

\def\linesolid{{------}}

\def\colr#1{\textcolor[rgb]{0.635,0.078,0.184}{#1}}  
\def\colo#1{\textcolor[rgb]{0.850,0.325,0.098}{#1}} 
\def\coly#1{\textcolor[rgb]{0.929,0.694,0.125}{#1}} 
\def\colg#1{\textcolor[rgb]{0.466,0.674,0.188}{#1}}  
\def\colb#1{\textcolor[rgb]{0.000,0.4470,0.7410}{#1}}    
   
\def\linesolid {{--------}}

\begin{document}
	
	\begin{frontmatter}
		
		
		\title{Multi-condition multi-objective optimization using deep reinforcement learning}
		
		\author{Sejin Kim \fnref{cont}}
		\author{Innyoung Kim \fnref{cont}}
		\author{Donghyun You\corref{cor1}}
		\address{Department of Mechanical Engineering, Pohang University of Science and Technology, 77 Cheongam-Ro, Nam-Gu, Pohang, Gyeongbuk 37673, South Korea \vspace{-0.4in}}
		
		\fntext[cont]{Authors contributed equally.}
		\cortext[cor1]{Corresponding author.}
		
		\ead{dhyou@postech.ac.kr}
		 
		\begin{abstract}
		A multi-condition multi-objective optimization method that can find Pareto front over a defined condition space is developed for the first time using deep reinforcement learning. Unlike the conventional methods which perform optimization at a single condition, the present method learns the correlations between conditions and optimal solutions. The exclusive capability of the developed method is examined in the solutions of a novel modified Kursawe benchmark problem and an airfoil shape optimization problem which include nonlinear characteristics which are difficult to resolve using conventional optimization methods. Pareto front with high resolution over a defined condition space is successfully determined in each problem. Compared with multiple operations of a single-condition optimization method for multiple conditions, the present multi-condition optimization method based on deep reinforcement learning shows a greatly accelerated search of Pareto front by reducing the number of required function evaluations. An analysis of aerodynamics performance of airfoils with optimally designed shapes confirms that multi-condition optimization is indispensable to avoid significant degradation of target performance for varying flow conditions.
		\end{abstract}
		
		
	\end{frontmatter}
	
	\section{Introduction}
    Optimization is central to all decision-making problems including economics, business administration, and engineering~\cite{chong2004introduction}. In particular, in applied mechanics such as structural mechanics, electromagnetism, and biomechanics, designing an optimal shape that maximizes target performance, so-called shape optimization, has been actively studied to this day~\cite{RN355,RN408,RN387,RN399}. Likewise, in fluid mechanics, studies on shape optimization for numerous real applications have been conducted~\cite{RN386}. For example, studies to improve aerodynamic or hydrodynamic characteristics through shape optimization of airplanes, ships, and automobiles have been continuing~\cite{RN390, RN388, RN409, RN381}. In addition, there have been many efforts in designing wind turbine blades that maximize power efficiency~\cite{RN396} and marine propellers to reduce underwater radiated noise~\cite{RN402}. Since fluids have nonlinear and high-dimensional characteristics, the performance of fluid machines can vary greatly depending on the shapes. Therefore, shape optimization is essential for the efficient operation of fluid machines.
	
	In practical applications, generally, the operating conditions and target performance of fluid machines vary depending on the situation. For example, in the case of wind turbines, varying wind conditions alter the aerodynamic and structural performance of the blades~\cite{RN397}. Also, in the case of aircraft, the aerodynamic requirements of the wing vary according to flight situations such as cruise, departure, and landing~\cite{RN382}. Therefore, in order to maintain the optimal state during operation, it is necessary to know the optimal shape under changing conditions and objectives and to modify its shape accordingly. In many application fields, numerous efforts have been made to maximize the target performance by changing the shape according to the situation~\cite{RN356}. For aircraft, many studies have been conducted to improve the performance by modifying the shape using smart adaptive devices and morphing materials~\cite{RN410, RN413, RN411}. In addition, morphing hydrofoils and morphing composite propellers have been actively studied in marine applications~\cite{RN389, RN403, RN401}. 
	
	Despite these efforts, ironically, related studies in the perspective of optimization methodology are insufficient. In order to remain optimal, knowing the optimal solution under changing conditions and objectives should be preceded. Optimization for various objectives is possible through conventional multi-objective (MO) optimization methods~\cite{RN391, RN392, RN393, RN357}. However, to the best of our knowledge, there is no optimization method that can find the optimal solution considering various conditions.	Therefore, in addition to MO optimization, an optimization method that can handle both various conditions and objectives is needed.
	
	MO optimization method is to optimize multiple objectives that generally conflict with each other at a single condition. The goal of MO optimization is to find Pareto front which is a set of optimal trade-off solutions among the objectives. However, it can be only applied to a single prescribed condition. To find optimal solutions within a condition range, it is necessary to prescribe some conditions as representatives in advance, and perform optimization separately. For example, Secanell \textit{et al.}~\cite{RN382} performed optimization at seven prescribed flight conditions to design a morphing airfoil. In addition, Wang \textit{et al.}~\cite{RN398} conducted optimization considering three representative conditions to design a centrifugal pump. However, if optimization is performed only at some prescribed conditions, the obtained solutions can be valid only at the predetermined conditions. In addition, because optimization has to be repeated from scratch for multiple conditions, it is very inefficient to perform optimization at sufficiently many conditions. Thus, in order to overcome these limitations, a multi-condition multi-objective (MCMO) optimization method that can efficiently find a set of optimal solutions (Pareto front) in a condition range is needed.
	
    Recently, with the advancement of artificial intelligence, studies combining it with optimization are being actively conducted~\cite{RN412, RN383}. In particular, deep reinforcement learning (DRL) is emerging as a new trend in the field of shape optimization~\cite{RN358}. Viquerat \textit{et al.}~\cite{RN400} showed the capability of DRL in shape optimization by successfully performing airfoil shape optimization at a single flow condition. Thereafter, DRL has started to be adopted for many shape optimization problems. Qin \textit{et al.}~\cite{RN385} conducted MO optimization of a cascade blade at a target flow condition using DRL. Also, Li \textit{et al.}~\cite{RN384} conducted airfoil shape optimization to reduce drag using DRL, and they showed that the learned network can extract more improved shapes compared to the original shape at unlearned conditions. As these studies indicate, DRL has ample potential to be a key to the development of a MCMO optimization method. The basic concept of DRL is to find an optimal action with a given state. Thus, if the condition and objective of optimization are set as the state, it can take a role of a MCMO optimizer. Moreover, in contrast to conventional methods where each condition should be treated independently, it is expected to be more efficient by learning the correlations between conditions and optimal solutions.
	
	In the present study, a DRL-based MCMO optimization method that can efficiently find Pareto front over a condition space is developed. Then, two MCMO optimization problems are dealt with by the developed method. The first problem is a benchmark problem to validate the method. As a benchmark problem, the Kursawe test function~\cite{Kursawe1990}, which is a representative MO optimization problem, is newly extended to be suitable for MCMO optimization. Next, it is applied to airfoil shape optimization to show its applicability in practical engineering applications. The airfoil shape optimization is a representative shape optimization problem involving fluid dynamics, where nonlinearity and high-dimensionality are combined. Despite these difficulties, it has been actively studied due to its direct applicability to numerous engineering fields~\cite{RN407, RN406, RN405}. Finally, further analysis is conducted in each problem to identify the exclusive capability of the proposed method.
	
	\section{Background} \label{sec2}
	\subsection{Multi-objective optimization} \label{sec2.1}
	\subsubsection{Problem description}\label{sec2.1.1}
	A MO optimization problem with the number of $m$ objective functions can be defined as follows:
	\begin{equation}
    \begin{gathered}
    \underset{\bf{x}}{\text{min}}\bf{f}(\bf{x}) = \underset{\bf{x}}{\text{min}}{(f_{1}(\bf{x}),\ldots,f_{m}(\bf{x})}) \\
    \text{subject to } \bf{x} \in \bf{\Omega},
    \end{gathered}
    \label{eq_mo}
    \end{equation}
    where $\bf{x}$ is an element in the decision space $\bf{\Omega}$, $\bf{f}$: $\bf{\Omega} \rightarrow \mathbb{R}^{m}$ consists of $m$ real-valued objective functions, and $\mathbb{R}^{m}$ is the objective space. 
	
	In this problem, generally, no single solution can optimize these objectives simultaneously as they conflict with each other. Instead, a set of optimal trade-off solutions among different objectives exists by the concepts of Pareto dominance and Pareto optimality, which are defined as follows:
	\begin{itemize}
	\item Pareto dominance: $\bf{x}$ is said to Pareto dominate $\bf{y}$, denoted by $\bf{x} \preceq \bf{y}$, if and only if $\forall i \in \{1,2,\ldots,m\}$, $f_{i}(\bf{x})\leq f_{i}(\bf{y})$ and $f_{j}(\bf{x}) < f_{j}(\bf{y})$ for at least one index $j \in \{1,2,\ldots,m\}$.
	\item Pareto optimality: A solution $\bf{x^{*}} \in \bf{\Omega}$ is said to be Pareto optimal if and only if $\nexists \bf{x} \in \bf{\Omega}$ such that $\bf{x} \preceq \bf{x^{*}}$.
	\end{itemize}
	The goal of a MO optimization problem is to find the Pareto optimal set and the corresponding Pareto front which is defined as $\{\bf{f}(\bf{x}) \in \mathbb{R}^{m} | \bf{x} \in $Pareto optimal set$\}$.

	\subsubsection{Weighted Chebyshev method}\label{sec2.1.2}
	The weighted Chebyshev method is one of decomposition-based methods for solving MO optimization problems. It scalarizes a MO optimization problem into multiple single-objective (SO) optimization problems by introducing a weight vector $\bf{w}$, and the Chebyshev scalarizing function $f_{Chebyshev}$. $\bf{w}$ determines the weight between objectives and $f_{Chebyshev}$ is the scalarized objective of each SO optimization problem. Then, the original MO optimization problem can be solved by performing a number of scalarized SO optimization processes with different $\bf{w}$. The scalarized SO optimization problem can be written as follows:
	\begin{equation}
	\begin{gathered}
	\underset{\bf{x}} {\text{min}}f_{Chebyshev}=\underset{\bf{x}} {\text{min}}(\underset{i=1,\ldots,m}{\text{max}} w_{i} |f_{i}(\bf{x}) - f_{i}^{*}| ) \\
	\text{where }|\bf{w}| = 1 \text{ and } w_{i} \geq 0.
	\end{gathered}
	\label{eq_cheby}
	\end{equation}
	$f_{i}^{*}$ is an utopia value which is defined as $f_{i}^{*} = \text{inf}\{{f_{i}(\bf{x})}\} - \tau_{i}$, where $\tau_{i} > 0$ is a relatively small value.
	
	Unlike other decomposition-based methods, the weighted Chebyshev method guarantees that all Pareto optimal solutions can be obtained for both convex and nonconvex problems~\cite{miettinen2012nonlinear}. Because of this advantage, it has been widely used in literature~\cite{RN394, RN404} and was also successfully used with DRL~\cite{RN395}. One of the difficulties in adopting the weighted Chebyshev method is that the utopia point, $\bf{f}^*$, has to be known before optimization, which requires SO optimization for each objective in advance. In the present study, the difficulty is overcome by integrating these overall processes into a single process, which will be further discussed in Section~\ref{sec4.1.1}.
	
	\subsection{Deep reinforcement learning based optimization} \label{sec2.2}
	\subsubsection{Deep reinforcement learning} \label{sec2.2.1}
	Reinforcement learning is a process of learning a policy to determine an optimal action with a given state~\cite{sutton2018reinforcement}. At each discrete step {\it n}, it determines an action according to its current policy $\bf{a}_n$ = $\pi(\bf{s}_n)$. Then, through the execution of the action, a reward $r_n$, according to the decision and the next state $\bf{s}_{n+1}$, are given. As the step progresses, data is accumulated and learning proceeds. The goal of learning is to find the optimal policy that maximizes the value function $Q^{\pi}(\bf{s}_n,\bf{a}_n)$~\cite{RN359}, which is defined as the expected sum of an immediate reward, $r_n$, and discounted future rewards as follows:
    \begin{equation}
        Q^{\pi}(\bf{s}_n,\bf{a}_n)={\mathop{\mathbb{E}}}(r_n+{\gamma}r_{n+1}+{\gamma}^2r_{n+2}+\ldots), 
        \label{eq_valuefunction}
    \end{equation}
    where $\gamma$ is a discount factor that determines the weight between short-term and long-term future rewards. This process is repeated until the terminal state, and it is called one episode. 
    
    In particular, if deep learning is adopted for learning, it is called DRL. For example, deep neural networks can be used as a policy itself or for predicting the value function. By incorporating the deep neural network, DRL is known to be able to handle complex and high-dimensional problems~\cite{RN360, mnih2013playing}. Especially, DRL has shown its outstanding ability in optimal control and optimization~\cite{RN358, RN361, RN362}. 
    
    \subsubsection{Single-step deep reinforcement learning based optimization} \label{sec2.2.2}
	Single-step DRL based optimization is very recently introduced by Viquerat \textit{et al.}~\cite{RN400} where one learning episode consists of a single step; if an action is determined with a given state, a reward is given accordingly and the episode ends without the next state. Since the future rewards in Eq.~\eqref{eq_valuefunction} do not exist, the discount factor, $\gamma$, does not have to be defined and learning proceeds to maximize only the immediate reward. As a result, the optimal action that maximizes the reward itself can be directly determined.	Therefore, if the reward is set as the objective function to be optimized, the optimal solution that maximizes the objective function can be directly obtained. By virtue of this characteristic, single-step DRL is known to be suitable as an optimization method~\cite{viquerat2021review}.
	
	\section{Problem description of multi-condition multi-objective optimization} \label{sec3}
	A MCMO optimization problem is extended from a MO optimization problem to include not only the decision variable $\bf{x}$, but also the condition variable $\bf{c}$. The problem with the number of $m$ objective functions is defined as follows: 
	\begin{equation}
    \begin{gathered}
    \underset{\bf{x}}{\text{min}}\bf{f}(\bf{x}, \bf{c}) = \underset{\bf{x}}{\text{min}}{(f_{1}(\bf{x}, \bf{c}),\ldots,f_{m}(\bf{x},\bf{c})}) \\
    \text{subject to } \bf{x} \in \bf{\Omega}, \bf{c} \in \bf{\Phi},
    \end{gathered}
    \label{eq_mcmo}
    \end{equation}
	where $\bf{x}$ is an element in the decision space $\bf{\Omega}$, $\bf{c}$ is an element in the condition space $\bf{\Phi}$, $\bf{f}$: $\bf{\Omega}\times\bf{\Phi} \rightarrow \mathbb{R}^{m}$ consists of $m$ real-valued objective functions, and $\mathbb{R}^{m}$ is the objective space.
	
	Likewise, the concepts of Pareto dominance and Pareto optimality are extended to cover the condition variable, $\bf{c}$, which are defined as follows:
	\begin{itemize}
	\item Pareto dominance: $\bf{x}$ is said to Pareto dominate $\bf{y}$ at the condition variable  $\bf{c} \in \bf{\Phi}$, denoted by $\bf{x} \preceq_{\bf{c}} \bf{y}$, if and only if $\forall i \in \{1,2,\ldots,m\}$, $f_{i}(\bf{x}, \bf{c})\leq f_{i}(\bf{y}, \bf{c})$ and $f_{j}(\bf{x}, \bf{c}) < f_{j}(\bf{y}, \bf{c})$ for at least one index $j \in \{1,2,\ldots,m\}$.
	\item Pareto optimality: A solution $\bf{x^{*}_{c}} \in \bf{\Omega}$ is said to be Pareto optimal at the condition variable $\bf{c} \in \bf{\Phi}$, if and only if $\nexists \bf{x} \in \bf{\Omega}$ such that $\bf{x} \preceq_{\bf{c}} \bf{x^{*}_{c}}$.
	\end{itemize}
	As in the MO optimization problem, solving a MCMO optimization problem is to find the Pareto optimal set and the corresponding Pareto front which is defined as $\{\bf{f}(\bf{x}_{\bf{c}}, \bf{c}) \in \mathbb{R}^{m} | \bf{x}_{\bf{c}} \in $Pareto optimal set$\}$. If $\bf{c}$ is fixed, the MCMO optimization problem is reduced to a MO optimization problem.
	 
	\section{Method} \label{sec4}
	
	\subsection{Deep reinforcement learning algorithm for multi-condition multi-objective optimization} \label{sec4.1}
	
	\subsubsection{State, action, and reward} \label{sec4.1.1}
	In MCMO optimization, the optimal solution varies depending on the condition and objective. Therefore, the state of DRL is set to include the condition and objective, which is defined as follows:
	\begin{equation}
    \bf{s} = [\bf{c},\bf{w},\bf{f}^*],
    \label{eq_state}
    \end{equation}
    where $\bf{c}$ is a condition variable, $\bf{w}$ is a weight vector, and $\bf{f}^*$ is a utopia point at that condition. In the present study, $\bf{f}^*$ is adaptively updated during optimization to a slightly lower value than the minimum value of each objective function. Since the Chebyshev scalarizing function, $f_{Chebyshev}$, differs depending on $\bf{f}^*$ as in Eq.~\eqref{eq_cheby}, the changing utopia information is included in the state for stable learning.
    
    
    The action of DRL determines $\bf{x}$, element in the decision space, according to its policy, which is defined as follows:
    \begin{equation}
    \bf{a} = [\bf{x}].
    \label{eq_action}
    \end{equation}
    In addition, all variables in the state and action are normalized to an absolute magnitude around $1$ for scaling.
    
    Lastly, the reward of DRL is a quantitative evaluation of an action, which is defined as follows:
    \begin{equation}
    r = -f_{Chebyshev} = -\underset{i=1,\ldots,m}{\text{max}} w_{i} |f_{i}(\bf{x}) - f_{i}^{*}|,
    \label{eq_reward}
    \end{equation}
    where $f_i(\bf{x})$ is the value of the objective function obtained by executing an action. Note that the minus sign is added because the aim of optimization is to find $\bf{x}$ minimizing the Chebyshev scalarizing function.
    
	\subsubsection{Data reproduction method} \label{sec4.1.2}
	In the present study, a data reproduction method is applied to enlarge the number and diversity of data by exploiting the nature of the Chebyshev scalarizing function, $f_{Chebyshev}$. As in Eq.~\eqref{eq_reward}, the reward of DRL is a function of $\bf{w}$, $\bf{f}$, and $\bf{f^{*}}$. Since $\bf{w}$ is independent of $\bf{f}$ and $\bf{f^{*}}$, different rewards can be determined for arbitrary $\bf{w}$, once the objective functions are evaluated. Therefore, it is possible to reproduce an original data of $(\bf{s}, \bf{a}, r)$ pair by changing $\bf{w}$ with a single function evaluation. It is expected that this method would accelerate learning and, thus, be essential for optimization problems where the function evaluation is costly. In the present study, at each episode, 100 data are reproduced from a single original data by changing $\bf{w}$ in a uniform distribution.
	
	\subsubsection{Learning procedure} \label{sec4.1.3}
	The learning procedure of the present study is summarized in Algorithm~\ref{alg_drl}. For the DRL algorithm, the actor-critic algorithm~\cite{konda2000actor} is used, which is one of the representative DRL algorithms. In the algorithm, two types of neural networks are introduced. One is an actor network, policy itself, which determines an action in continuous space. The other is a critic network which predicts the value function depending on the state and action. As learning progresses, the critic network predicts the value function more and more accurately and, based on this, the probability that the actor network selects the optimal action increases.
	
    Both networks are set as fully connected networks with four hidden layers of $512$, $256$, $256$, and $128$ neurons and the Leaky ReLU activation function~\cite{maas2013rectifier} is used for the hidden layers in both networks. At the output layer of the actor network, the Tanh activation function is added so that the action values range from $-1$ to $1$. The learning rates are equally set to $10^{-4}$ and Adam optimizer~\cite{kingma2017adam} is used for updating the network parameters. Especially, the actor network is updated every two learning iterations ($l_d=2$) for stable learning~\cite{RN363}. The mini-batch size, $N_b$, is set to $100$ and the learning amount per one episode, $l_{max}$, is set to $100$ which is the same as the number of reproduced data per one original data. The standard deviation of the exploration noise, $\sigma$, is set to $1$ in the initial $1000$ warm-up episodes and $0.05(\text{cos}(2\pi/1000\times\text{episode})+1)$ afterward. The use of the cosine function enables both exploration for avoiding local minima and exploitation for accurately finding optimal solutions.
    
	\subsection{Selection of Pareto front} \label{sec4.2}
	Pareto dominance in a MCMO optimization problem is defined at each condition variable $\bf{c} \in \bf{\Phi}$. However, as $\bf{f(\bf{x},\bf{c})}$ in the obtained data are scattered over $\bf{\Phi}$, there is no exactly the same $\bf{c}$ where the dominance can be judged. Therefore, a concept of decomposition of the condition space is introduced to derive approximate solutions for a MCMO optimization problem. $\bf{\Phi}$ is decomposed into $N$ spaces as follows:
	\begin{equation}
    \begin{gathered}
    \bf{\Phi}_{1} \cup \bf{\Phi}_{2} \cup \ldots \cup \bf{\Phi}_{N} = \bf{\Phi}, \\
    \bf{\Phi}_{i} \cap \bf{\Phi}_{j} = \varnothing \text{ where } i \neq j.
    \end{gathered}
    \label{eq_n_grid}
    \end{equation}
	In each $\bf{\Phi}_{i}$, $\bf{c}$ is assumed to be the same, and Pareto front is selected from the data.
	
	Note that the decomposition has no effect on the optimization process and can be modified during or after the optimization process. Therefore, $N$ can be freely adjusted according to the desired quality. For example, the denser the decomposition, the higher the resolution of the selected Pareto front, but the number of episodes required for convergence increases. In the present study, $\bf{\Phi}$ is decomposed into 100 spaces of the same size for selecting Pareto front.

	\subsection{Convergence judgment} \label{sec4.3}
\begin{figure}
	\centering
	\includegraphics[width=0.6\linewidth]{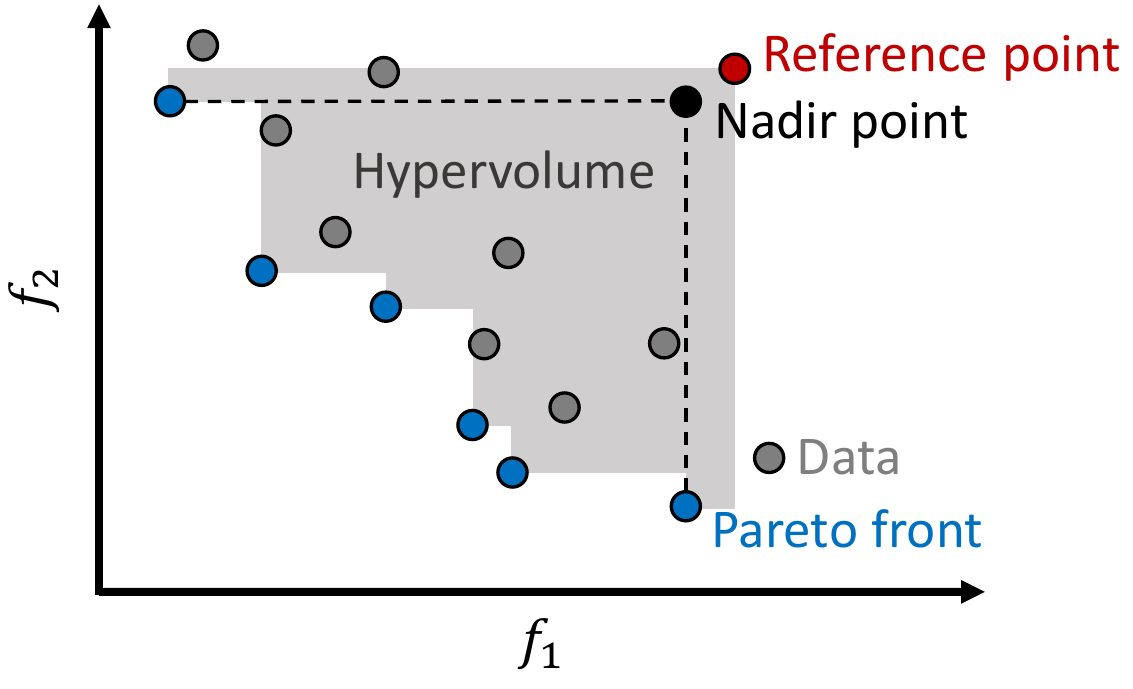}
	\caption{Schematic of the HV when the number of objectives is two.} \label{fig_HV}
	\end{figure}

	In order to judge the convergence of an optimization process, the hypervolume indicator (HV)~\cite{zitzler1999evolutionary} is adopted. It refers to the volume in the objective space between Pareto front and a fixed reference point as shown in Fig.~\ref{fig_HV}. Due to its monotonic characteristic, the larger the HV, the more accurate the Pareto front. Therefore, it is one of the most frequently used indicators for convergence and capability assessment of MO optimization methods. A general guideline for determining the reference point is to use a slightly worse point than the nadir point consisting of the worst objectives values over the Pareto front~\cite{RN364}.
	
	In MCMO optimization, as described in Section~\ref{sec4.2}, the condition space, $\bf{\Phi}$, is decomposed and Pareto front is selected respectively in each decomposed space. Likewise, the HV is defined in each decomposed space. Therefore, in this study, the convergence of an optimization process is judged by HV$_{avg}$, the average HV over all the decomposed spaces.
	
	\section{Results and discussion} \label{sec5}
	In this section, the proposed DRL-based MCMO optimization method is applied to two problems, and the results are analyzed. The first problem is a newly modified Kursawe test function to a MCMO optimization problem. The second problem is airfoil shape optimization which is a representative shape optimization problem involving fluid dynamics.
	
	
	\subsection{Modified Kursawe test function} \label{sec5.1}
	
	\subsubsection{Problem setup} \label{sec5.1.1}
	The modified Kursawe problem for MCMO optimization is defined as follows:
    \begin{equation}
    \begin{gathered}
    \underset{\bf{x}}{\text{min}}\bf{f}(\bf{x},\bf{c}) = \underset{\bf{x}}{\text{min}}{(f_{1}(\bf{x},\bf{c}),f_{2}(\bf{x},\bf{c})}) \\
    \text{subject to } \bf{x} \in \bf{\Omega}, \bf{c} \in \bf{\Phi}, \\
    \text{where } f_{1}(\bf{x},\bf{c}) = g_{1}(\bf{x})\text{cos}\theta_r - g_{2}(\bf{x})\text{sin}\theta_r, \\
    f_{2}(\bf{x},\bf{c}) = g_{1}(\bf{x})\text{sin}\theta_r + g_{2}(\bf{x})\text{cos}\theta_r, \\
    g_{1}(\bf{x}) = \sum_{i=1}^{2} (-10\text{exp}(-0.2 \sqrt{x_i^2+x_{i+1}^2})), \\
    g_{2}(\bf{x}) = \sum_{i=1}^{3} (|x_i|^{0.8}+5\text{sin}(x_i^3)), \\ 
    \bf{\Omega} = \{(x_{1}, x_{2}, x_{3}) | -5 \leq x_{i} \leq 5,~i=1,2,3\}, \\
    \bf{\Phi} = \{(\theta_r) | 0 \leq \theta_r \leq \pi/4\}.
    \end{gathered}
    \label{eq_kursawe_3d}
    \end{equation}
    
    For the modification, $\theta_r$ is introduced for rotational transformation. If $\theta_r$ is set to $0$, it reduces to the original Kursawe problem which is a MO optimization problem. Extending the problem through rotational transformation has two advantages. First, the characteristics of the original problem can be preserved. As the original Kursawe problem has discontinuous and nonconvex Pareto front, it has been actively adopted to evaluate the capability of MO optimization methods~\cite{RN365, RN367, RN368, RN369}. Thus, with the preserved characteristics, the modified Kursawe problem can be a satisfactory benchmark problem for validating the developed MCMO optimization method. Second, real solutions can be readily obtained, which is crucial in designing a benchmark problem. The boundary shape of the feasible region in the objective space remains unchanged from the original Kursawe problem by rotational transformation. Therefore, the real Pareto front corresponding to $\theta_r$ can be easily obtained by judging dominance from the rotated boundary.
	
	\subsubsection{Optimization results} \label{sec5.1.2}
\begin{figure}[]
	\begin{subfigure}[t]{\linewidth}
		\includegraphics[width=\textwidth]{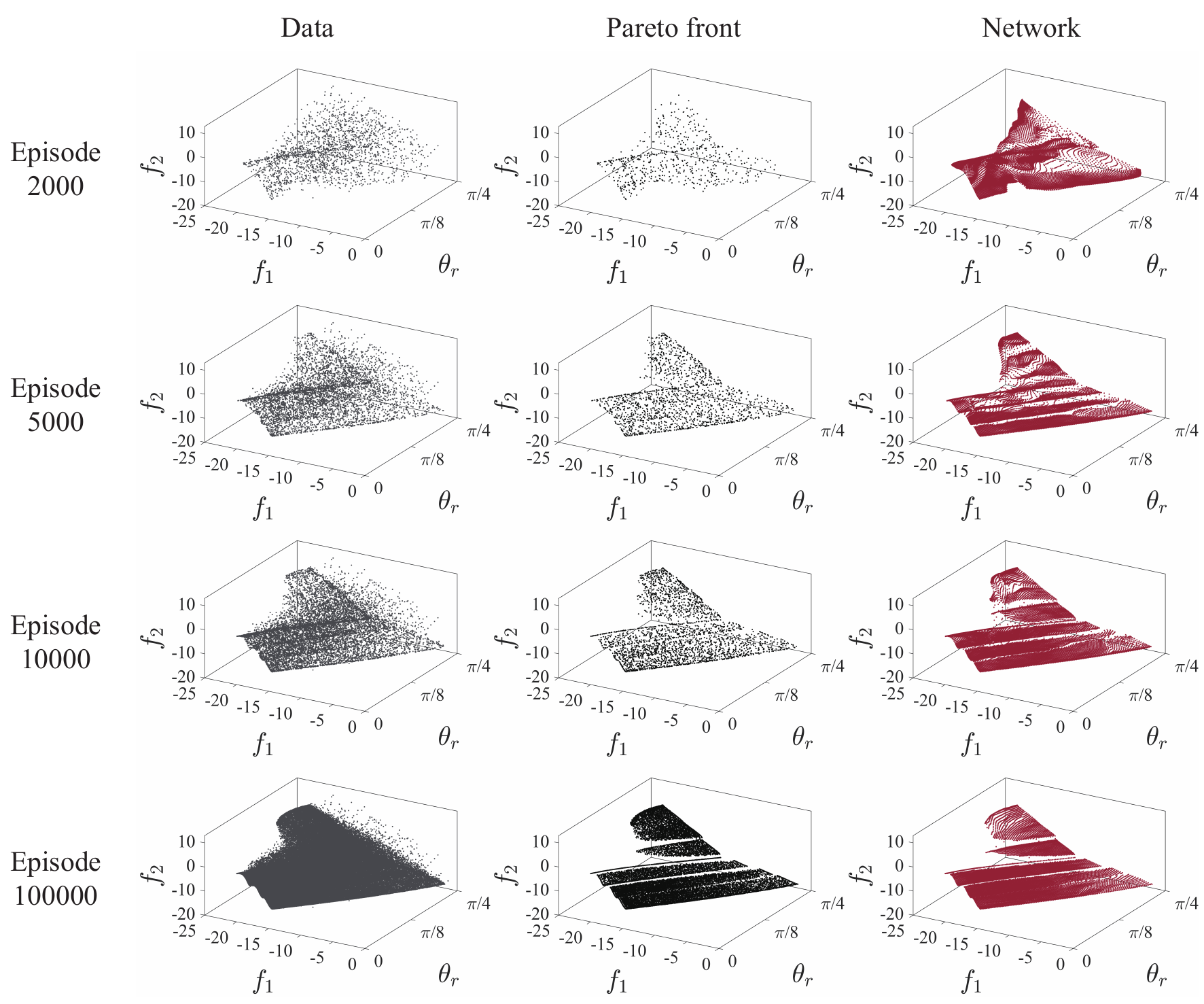}
		\caption{Results during the optimization process. One data point is added for every episode. The Pareto front is selected from the data as described in Section~\ref{sec4.2}. The network shows the values of objective functions corresponding to the actions determined by the network.} \label{fig_kursawe_3d_process_a}
	\end{subfigure}
	\begin{subfigure}[t]{\linewidth}
        \centering
		\includegraphics[width=0.5\textwidth]{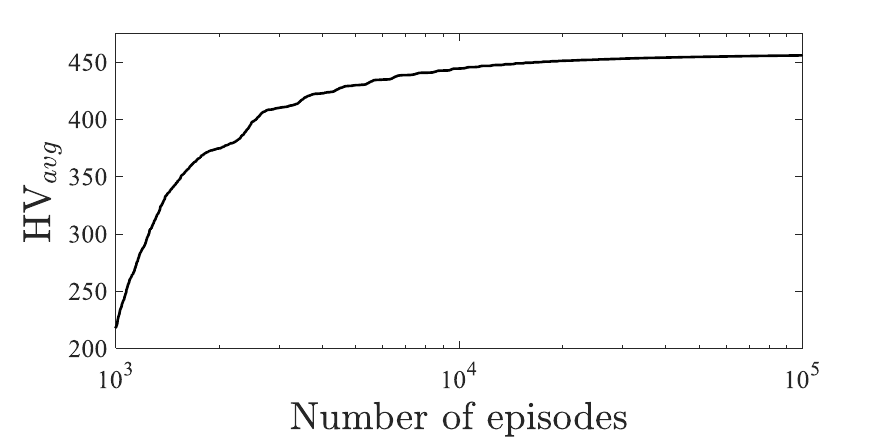}
		\caption{$\text{HV}_{avg}$ over episodes with a reference point $(f_1^{ref},f_2^{ref})=(-2,13)$.} \label{fig_kursawe_3d_process_b}
	\end{subfigure}
	\caption{Optimization process of the modified Kursawe test function.} \label{fig_kursawe_3d_process}
	\end{figure}

	The modified Kursawe problem is solved as described in Algorithm~\ref{alg_drl}. Fig.~\ref{fig_kursawe_3d_process} shows the optimization process. As shown in Fig.~\ref{fig_kursawe_3d_process_a}, in the early episodes, data are widely scattered as the network is not developed enough. However, as the episode progresses, data are accumulated, and the network learns to find an optimal action with a given state. As a result, better solutions are obtained for newly given conditions and objectives, increasing the resolution of the Pareto front. Also, the clustered data near the Pareto front reinforce the learning again, forming a positive feedback loop. Finally, the Pareto front and the network converge at episode $100000$, which can be also seen in terms of HV$_{avg}$ as shown in Fig.~\ref{fig_kursawe_3d_process_b}.
	
\begin{figure}[]
	\centering
	\begin{subfigure}[t]{\linewidth}
		\includegraphics[width=\textwidth]{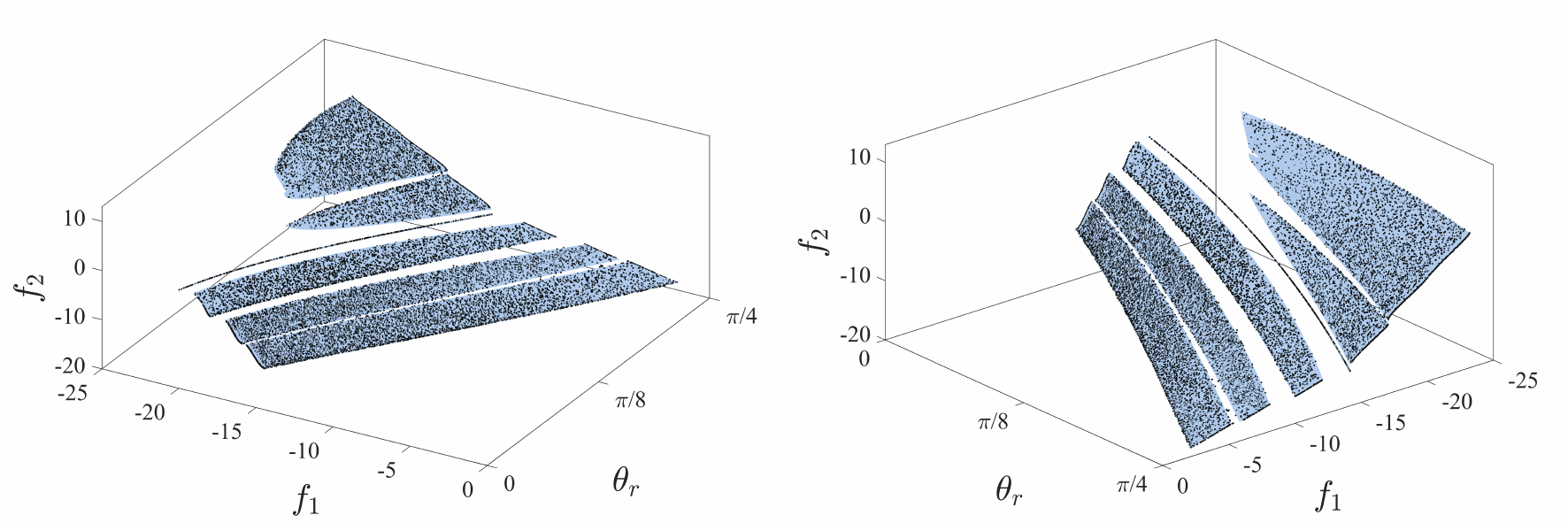}
		\caption{Pareto front from two different angles.} \label{fig_kursawe_3d_results_a}
	\end{subfigure}
	\begin{subfigure}[t]{\linewidth}
		\includegraphics[width=\textwidth]{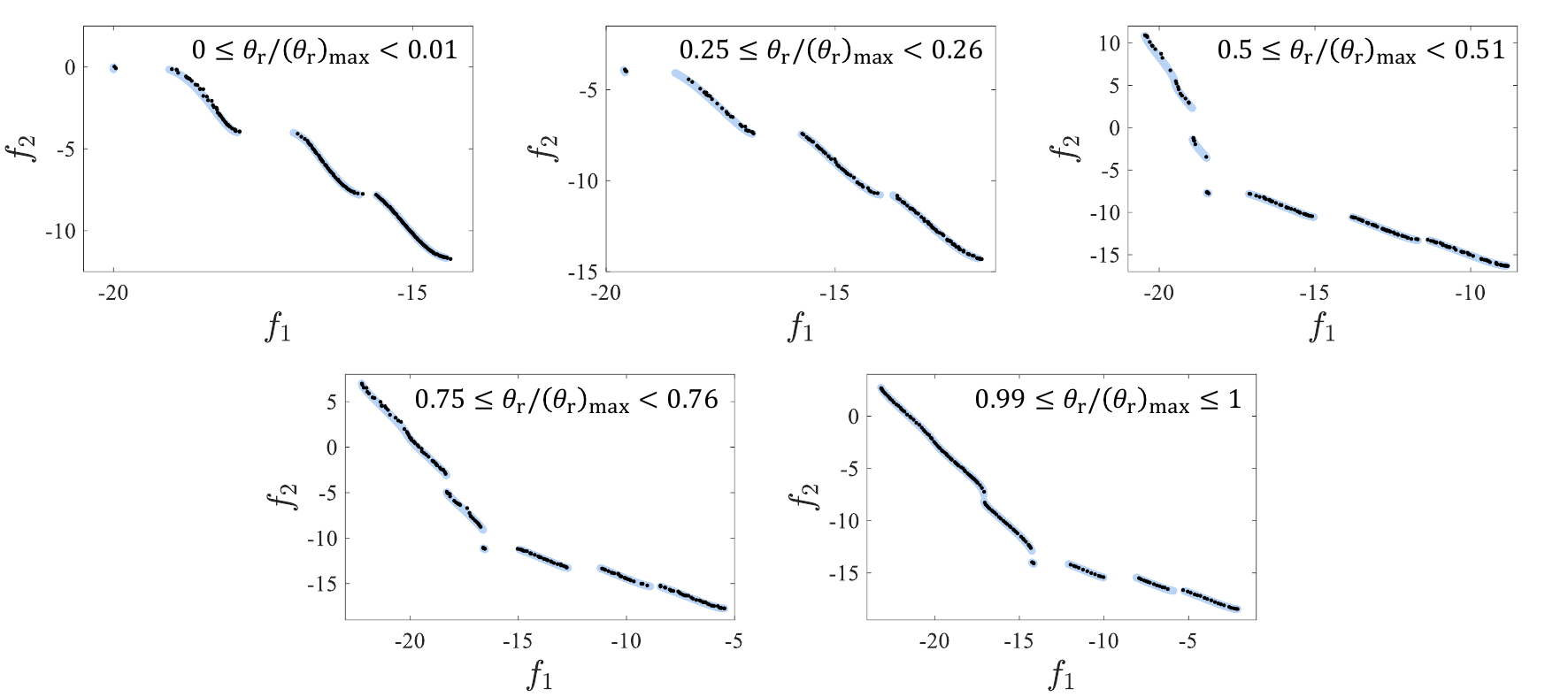}
		\caption{Pareto front in five decomposed condition spaces} \label{fig_kursawe_3d_results_b}
	\end{subfigure}
	\caption{Optimization results of the modified Kursawe test function. The black dots show the Pareto front at episode 100000. The sky-blue surfaces in (a) and lines in (b) show real Pareto front depicted for comparison.} \label{fig_kursawe_3d_results}
	\end{figure}

	Fig.~\ref{fig_kursawe_3d_results} shows the optimization results at the converged episode. Through the optimization, high-resolution Pareto front of $15268$ solutions over the whole condition space is obtained. As shown in Fig.~\ref{fig_kursawe_3d_results_a}, it shows good agreement with the real Pareto front including highly nonlinear parts near $\theta_r/(\theta_r)_{max}=0.4$ where the shape of the Pareto front drastically changes along $\theta_r$. It shows the exclusive ability of the proposed MCMO optimization method. If several representative conditions are predetermined and optimization is performed at each condition, it is difficult to capture the nonlinear parts. On the other hand, the developed MCMO optimization is performed over the entire condition space, so that high-resolution Pareto front can be found. Fig.~\ref{fig_kursawe_3d_results_b} shows the optimization results in five decomposed condition spaces. Note that since the condition space, $\bf{\Phi}$, is equally decomposed into $100$ spaces in the present study, each figure in Fig.~\ref{fig_kursawe_3d_results_b} shows one decomposed condition space. Even in those decomposed spaces, the solutions match well with the real Pareto front. 
	
	\subsubsection{Effectiveness of multi-condition optimization} \label{sec5.1.3}
	In this section, a computational experiment based on the modified Kursawe problem is set to analyze how effective multi-condition (MC) optimization is compared to single-condition (SC) optimization. The experiment is designed to compare the number of function evaluations to reach the same quality of optimization. Since SC optimization cannot be performed over a condition space, equally distributed $N_c$ conditions are prescribed in the condition space, $\bf{\Phi}$, for the comparison. To measure the quality of optimization, a reference $\text{HV}_{i}^{ref}$ is set at each prescribed $i^{th}$ condition. 
	
	Then, two cases are compared by the total number of function evaluations required to reach the same $\textbf{HV}^{ref}$ at all prescribed conditions. The first case is SC optimization performed independently at each prescribed condition. The SC optimization method can be easily derived by fixing the condition in the developed method. The second case is MC optimization modified to be conducted only at the prescribed conditions. Although the proposed method in this study is conducted over a whole condition space, it is modified in this experiment for fair comparison.
	
\begin{figure}[]
	\begin{subfigure}[t]{\linewidth}
		\includegraphics[width=\textwidth]{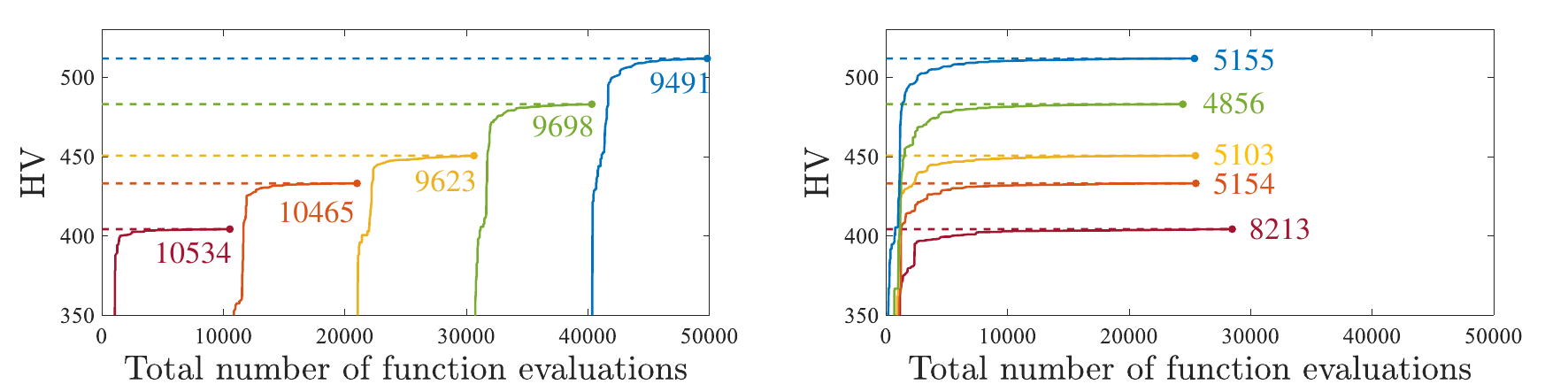}
		\caption{The HVs over the total number of function evaluations with a reference point $(f_1^{ref},f_2^{ref})=(-2,13)$ at five prescribed conditions. $\theta_r/(\theta_r)_{max}=$ \colr{\linesolid}, $0/4$; \colo{\linesolid}, $1/4$; \coly{\linesolid}, $2/4$; \colg{\linesolid}, $3/4$; \colb{\linesolid}, $4/4$. The left figure shows SC optimization and the right figure shows MC optimization. The dashed lines in both figures indicate the same $\textbf{HV}^{ref}$. If the HV at a specific condition satisfies $\textbf{HV}^{ref}$, the condition is excluded. The numbers near the lines are the numbers of function evaluations at each condition.} \label{fig_kursawe_exp_nc5_a}
	\end{subfigure}
	\begin{subfigure}[t]{\linewidth}
	    \centering
		\includegraphics[width=0.5\textwidth]{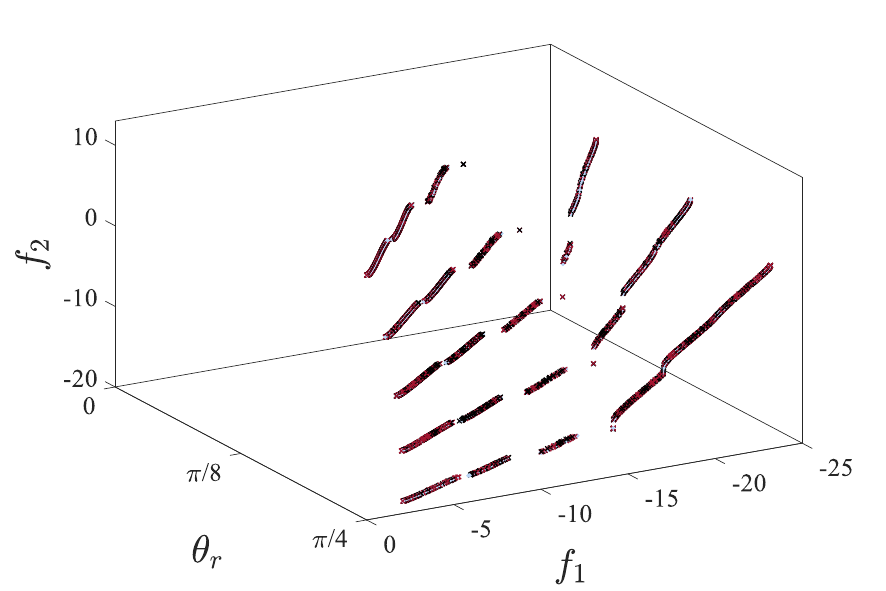}
		\caption{Pareto front obtained from the two cases. The black crosses show the Pareto front obtained form SC optimization and the red crosses show the Pareto front obtained from MC optimization. The sky-blue lines show the real Pareto front at the five prescribed conditions.} \label{fig_kursawe_exp_nc5_b}
	\end{subfigure}
	\caption{One of the experiment results when $N_c=5$.} \label{fig_kursawe_exp_nc5}
	\end{figure}

	$\textbf{HV}^{ref}$ at each condition is determined as an average HV by performing SC optimization ten times up to episode $10000$ since the optimization process is stochastic due to the exploration of DRL. The HV of SC optimization at each condition shows convergence around episode $10000$ and the obtained Pareto front matches well with the real Pareto front as shown in Fig.~\ref{fig_kursawe_exp_nc5}. This is quite comparable to other studies using the Kursawe test function for evaluating their optimization methods~\cite{RN365, RN367, RN368, RN369}. In addition, when comparing the two cases, the average number of total function evaluations of ten runs is used for precise analysis.
	
	Fig.~\ref{fig_kursawe_exp_nc5} shows one example of the results when $N_c=5$. As shown in Fig.~\ref{fig_kursawe_exp_nc5_a}, SC optimization is performed independently at each prescribed condition while MC optimization is performed simultaneously at the five prescribed conditions. As shown in the figure, the number of function evaluations at each condition is reduced in the MC optimization, resulting in a significant reduction of the total number of function evaluations. Total 28481 function evaluations are required in the SC optimization while total 49811 function evaluations are required to reach the same $\textbf{HV}^{ref}$ in the MC optimization. This reduction is attributed to the fact that the MC optimization learns the correlations between the conditions and the optimal solutions. By utilizing the correlations, it can effectively find the Pareto front with a small number of function evaluations. Fig.~\ref{fig_kursawe_exp_nc5_b} shows the Pareto front obtained from the two cases. Because both cases satisfy the same $\textbf{HV}^{ref}$, the Pareto front shows good agreement with the real Pareto front in both cases.
	
\begin{figure}[]
	\centering
    \includegraphics[width=0.75\textwidth]{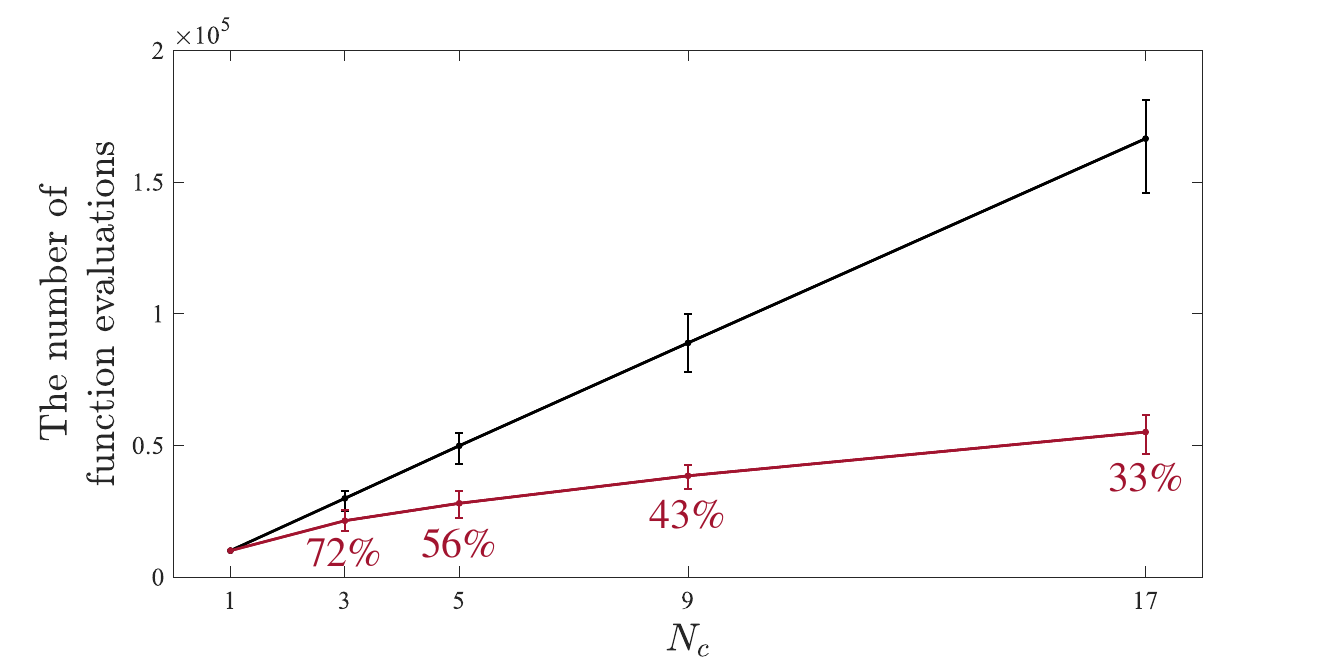}
	\caption{The number of function evaluations required to reach the same $\textbf{HV}^{ref}$ according to $N_c$. The black line indicates the average of SC optimization and the red line indicates the average of MC optimization. The error bars show the minimum and maximum values over ten runs. The numbers below the red line are the percentages of MC optimization compared to SC optimization.} \label{fig_kursawe_exp_improvement}
	\end{figure}

	Fig.~\ref{fig_kursawe_exp_improvement} shows the experiment results according to $N_c$. In SC optimization, the number of function evaluations increases linearly with $N_c$. This is a natural result because optimization is performed at each condition independently. However, in MC optimization, the increment gradually decreases, so that the difference between the two cases increases with $N_c$. Especially, when $N_c=17$, the number of function evaluations of MC optimization is only $33\%$ of that of SC optimization. Considering the proposed method in the present study is conducted continuously over a whole condition space ($N_c \rightarrow \infty$), it can be inferred that the reduction of the number of required function evaluations will be much greater than this result. Therefore, we can conclude that MC optimization is much effective than SC optimization and it is enabled by learning the correlations between conditions and optimal solutions.
	
	\subsection{Airfoil shape optimization} \label{sec5.2}
	\subsubsection{Problem setup} \label{sec5.2.1}
	In numerous engineering fields, the flow condition and aerodynamic requirement of an airfoil can vary depending on the situation. In this section, a MCMO airfoil shape optimization problem is defined reflecting the practical applications. First, the lift coefficient, $C_{L}$, and the lift-to-drag ratio, $C_{L}/C_{D}$, are set as the objectives of optimization to be maximized. These objectives are crucial factors in designing an airfoil which many researchers are interested in~\cite{RN370, RN371, RN372, huyse2002probabilistic}. Next, as a representative value of the flow condition, the chord Reynolds number, $Re_{c}$, is set as a condition variable of optimization.
	
	In this study, an airfoil shape is parameterized using the Kármán-Trefftz transformation~\cite{milne1973theoretical}. A Kármán-Trefftz airfoil is generated from the transformation of a circle in the $\zeta$-plane to the physical $z$-plane. The circle in the $\zeta$-plane centered on $(\mu_x,\mu_y)$ is defined to pass $(1,0)$. Then, a complex variable $\zeta = \xi+i\eta$ on the circle is transformed to $z= x +iy$ to generate an airfoil as follows:
	\begin{equation}
    \begin{gathered}
    z = n \frac{(\zeta+1)^{n} + (\zeta-1)^{n}}{(\zeta+1)^{n} - (\zeta-1)^{n}}, \\
    n = 2 - \frac{\beta}{180^{\circ}},
    \end{gathered}
    \label{eq_KT}
    \end{equation}
	where $\beta$ is a trailing-edge angle of the generated airfoil. Since it can generate various and realistic airfoils, it has been utilized in many studies~\cite{Paola2012, RN373}. Along with the shape of an airfoil itself, the angle of attack $\alpha$ is an important factor that greatly influences the aerodynamic characteristics. Thus, when designing an airfoil, $\alpha$ relative to the flow direction has to be optimized to achieve optimal performance~\cite{huyse2002probabilistic}. In the present study, in addition to $\mu_{x}$, $\mu_{y}$, and $\beta$ which determine a Kármán-Trefftz airfoil, $\alpha$ is included as a design variable of optimization.
    
    The MCMO airfoil shape optimization problem is defined as follows:
	\begin{equation}
    \begin{gathered}
    \underset{\bf{x}}{\text{min}}\bf{f}(\bf{x}, \bf{c}) = \underset{\bf{x}}{\text{min}}{(f_{1}(\bf{x}, \bf{c}), f_{2}(\bf{x},\bf{c})}) \\
    \text{subject to } \bf{x} \in \bf{\Omega}, \bf{c} \in \bf{\Phi}, \\ 
    \text{where } f_{1}(\bf{x}, \bf{c}) = -1/100 ~ C_{L}/C_{D}, \\
    f_{2}(\bf{x}, \bf{c}) = -C_{L}, \\ 
    \bf{\Omega} = \{(\mu_{x}, \mu_{y}, \beta, \alpha) | -0.4 \leq \mu_{x} \leq -0.05, ~ 0 \leq \mu_{y} \leq 0.4, ~ 1 ^{\circ} \leq \beta \leq 30 ^{\circ}, ~ 0 ^{\circ} \leq \alpha \leq 30 ^{\circ}\}, \\
	\bf{\Phi} = \{(Re_{c}) | 10^{5} \leq Re_{c} \leq 10^{7}\}.
    \end{gathered}
    \label{eq_airfoil_MCMOO}
    \end{equation}
    Since the goal of optimization is to maximize $C_{L}/C_{D}$ and $C_{L}$, minus signs are added. $1/100$ is multiplied to $C_{L}/C_{D}$ to match the scale between $f_{1}(\bf{x}, \bf{c})$ and $f_{2}(\bf{x}, \bf{c})$. In order to evaluate $C_{L}/C_{D}$ and $C_{L}$, XFOIL which is an analysis tool for airfoils~\cite{RN374} is adopted in the present study. It is widely used for airfoil shape optimization due to its low computational cost~\cite{RN375, RN376, RN377}. The ranges of the design variables, $\mu_x$, $\mu_y$, $\beta$, and $\alpha$, are set to generate various airfoil shapes excluding unrealistic shapes, and the range of $Re_{c}$ is set to cover sufficiently wide applications~\cite{RN378}.
	
	\subsubsection{Optimization results} \label{sec5.2.2}
\begin{figure}[]
	\begin{subfigure}[t]{\linewidth}
	    \centering
		\includegraphics[width=0.5\textwidth]{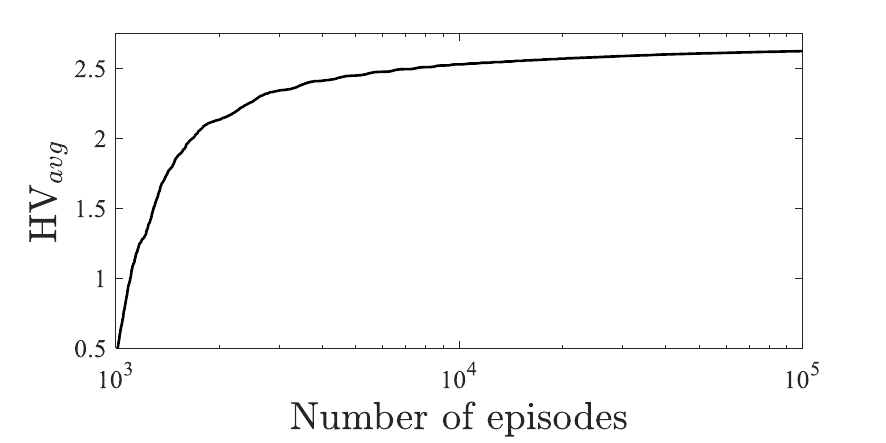}
		\caption{$\text{HV}_{avg}$ over episodes with a reference point $(f_1^{ref},f_2^{ref})=(0,1)$.} \label{fig_airfoil_3d_results_a}
	\end{subfigure}
	\begin{subfigure}[t]{\linewidth}
		\includegraphics[width=\textwidth]{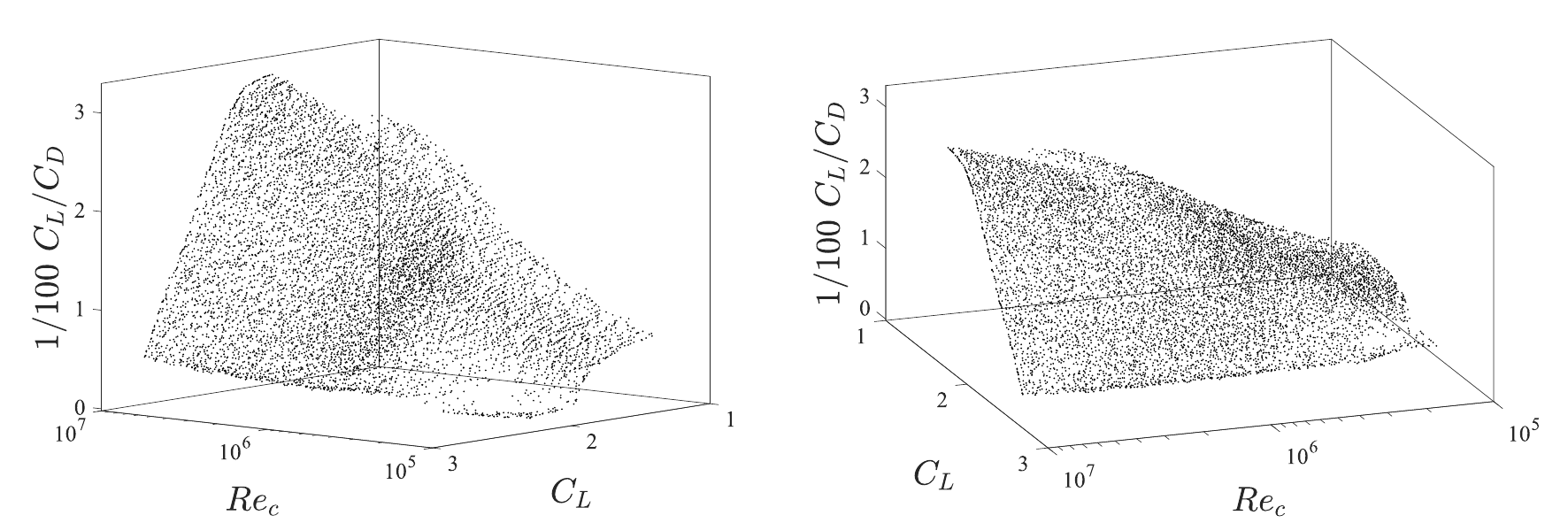}
		\caption{Pareto front at episode $100000$ from two different angles.} \label{fig_airfoil_3d_results_b}
	\end{subfigure}
	\caption{Results of airfoil shape optimization.} \label{fig_airfoil_3d_results}
	\end{figure}

	The airfoil shape optimization problem is solved as described in Algorithm~\ref{alg_drl}. Fig.~\ref{fig_airfoil_3d_results} shows the results of airfoil shape optimization. As shown in Fig.~\ref{fig_airfoil_3d_results_a}, $\text{HV}_{avg}$ is shown to converge at episode $100000$. Fig.~\ref{fig_airfoil_3d_results_b} shows the Pareto front at the converged episode. Overall, Pareto front of sufficient resolution is successfully found within the defined condition space, indicating that the developed method can be applied to practical engineering applications. As shown in Fig~\ref{fig_airfoil_3d_results_b}, the maximum $C_{L}/C_{D}$ increases with $Re_{c}$ while the maximum $C_{L}$ remains relatively constant. In particular, along the line where $C_{L}/C_{D}$ is maximized, $C_{L}$ does not change significantly, which refers that $C_{D}$ decreases according to $Re_{c}$. In addition, two distinct features are observed in the Pareto front. When $C_{L}$ is maximized, nonconvex parts are observed near $Re_{c}=10^{5}$. Next, when $C_{L}/C_{D}$ is maximized, nonlinear parts are observed near $Re_{c}=2\times10^{6}$. These parts will be further discussed through analysis of the optimal solutions and the optimal airfoil shapes.
	
\begin{figure}[]
	\begin{subfigure}[t]{\linewidth}
	    \centering
		\includegraphics[width=0.8333\textwidth]{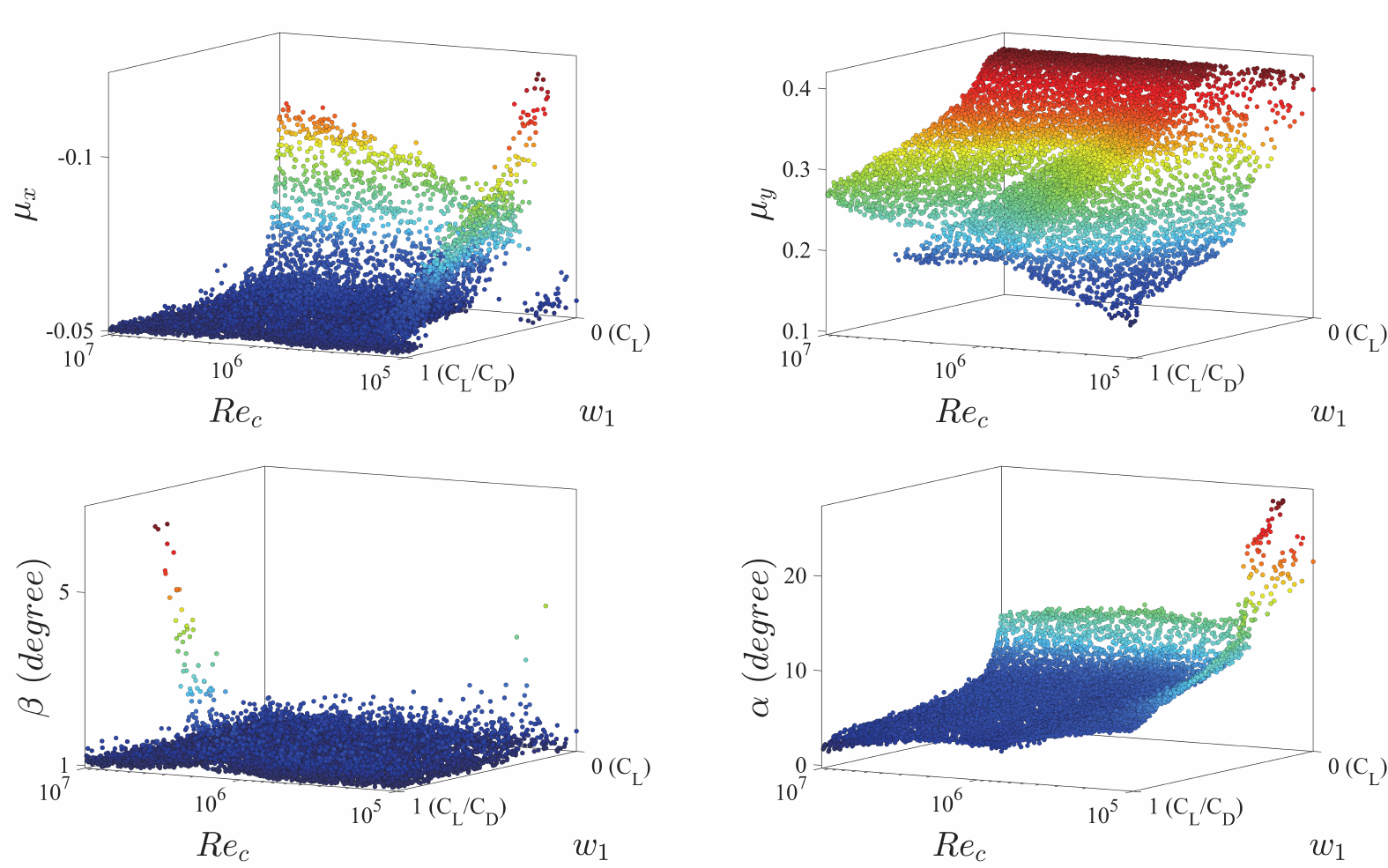}
		\caption{Optimal design parameters corresponding to $Re_{c}$ and $w_{1}$. $w_{1}$ is the weight of $C_{L}/C_{D}$. Thereby, $w_{1} = 1$ corresponds to the maximization of $C_{L}/C_{D}$ while $w_{1} = 0$ corresponds to the maximization of $C_{L}$.} \label{fig_airfoil_3d_results2_a}
	\end{subfigure}
	\begin{subfigure}[t]{\linewidth}
		\includegraphics[width=\textwidth]{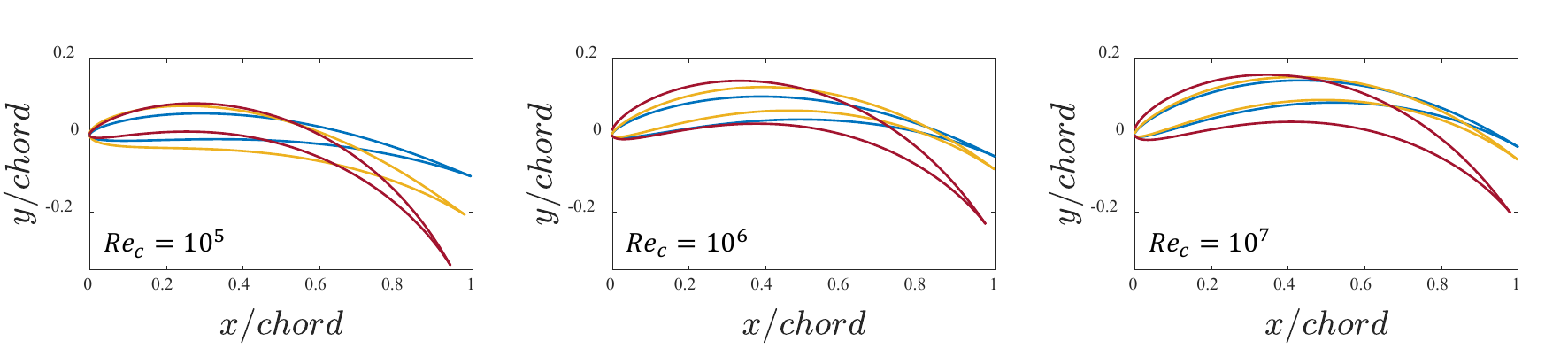}
		\caption{Optimal airfoil shapes at different $Re_{c}$ and $w_{1}$.
		$w_{1}=$ \colr{\linesolid}, $0$; \coly{\linesolid}, $0.5$; \colb{\linesolid}, $1$. $\alpha$ of the airfoils are expressed assuming that the flow direction is horizontal.} \label{fig_airfoil_3d_results2_b}
	\end{subfigure}
	\caption{Optimal solutions and optimal airfoil shapes.} \label{fig_airfoil_3d_results2}
	\end{figure}

	Fig.~\ref{fig_airfoil_3d_results2} shows the optimal solutions and the optimal airfoil shapes. As can be seen in Fig.~\ref{fig_airfoil_3d_results2_a}, various values of design parameters are obtained depending on $Re_{c}$ and $w_1$. For each design parameter, $\mu_{x}$ is a factor that determines the thickness of the airfoil, so the smaller the absolute value, the thinner airfoil is generated. $\mu_{y}$ determines the camber of the airfoil. $\mu_{y}=0$ indicates a symmetric airfoil, and the larger the value, the upper cambered airfoil is generated. $\beta$ and $\alpha$ are the trailing-edge angle and the angle of attack respectively, which are expressed in degrees.
	
	As shown in Fig.~\ref{fig_airfoil_3d_results2_a}, nonlinear features are observed where the optimal design parameters change dramatically with respect to $Re_{c}$ and $w_{1}$. These are particularly noticeable for $\mu_{x}$ and $\alpha$ near $Re_{c} = 10^{5}$ and $w_{1} = 0$, and for $\mu_{y}$ and $\beta$ near $Re_{c} = 2\times10^{6}$ and $w_{1} = 1$. These parts correspond to the aforementioned nonconvex and nonlinear parts observed in Pareto front respectively. Except for these nonlinear parts, overall trends are observed. As the weight of $C_{L}/C_{D}$ increases, thin and less cambered airfoils with low $\alpha$ are generated. On the other hand, As the weight of $C_{L}$ increases, thick and highly cambered airfoils with high $\alpha$ are generated. The trailing-edge angle, $\beta$, shows relatively less variation and keeps its minimum value.
	
	Fig.~\ref{fig_airfoil_3d_results2_b} is the optimal airfoil shapes according to $Re_{c}$ and $w_{1}$. As mentioned above, as $w_{1}$ is close to $0$, airfoils with high camber and $\alpha$ are generated for maximizing the lift at all $Re_{c}$. However, the values of $\alpha$ do not increase to the maximum, which is due to the consideration of a stall phenomenon caused by excessively high $\alpha$. On the contrary, as $w_{1}$ is close to $1$, the opposite tendency is observed to consider the drag. Also, when considering the drag, the thicknesses of airfoils decrease except for $Re_{c} = 10^{5}$ where the aforementioned nonlinearity exists.
	
	\subsubsection{Aerodynamic performance analysis of optimal airfoil shapes} \label{sec5.2.3}
\begin{figure}[]
	\centering
	\begin{subfigure}[t]{\linewidth}
		\includegraphics[width=\textwidth]{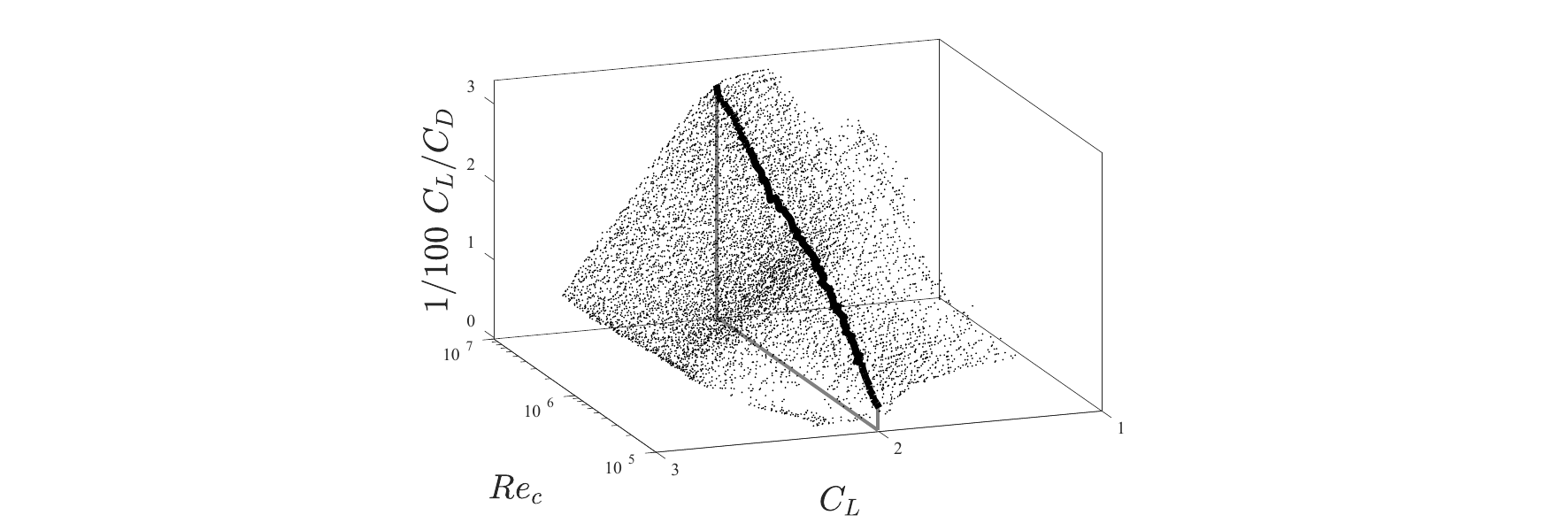}
		\caption{Finding the optimal shapes that maximize $C_{L}/C_{D}$ with a constraint $C_{L} \geq 2$ in Pareto front. The black line corresponds to the optimal shapes.} \label{fig_airfoil_3d_analysis_a}
	\end{subfigure}
		\begin{subfigure}[t]{\linewidth}
		\includegraphics[width=\textwidth]{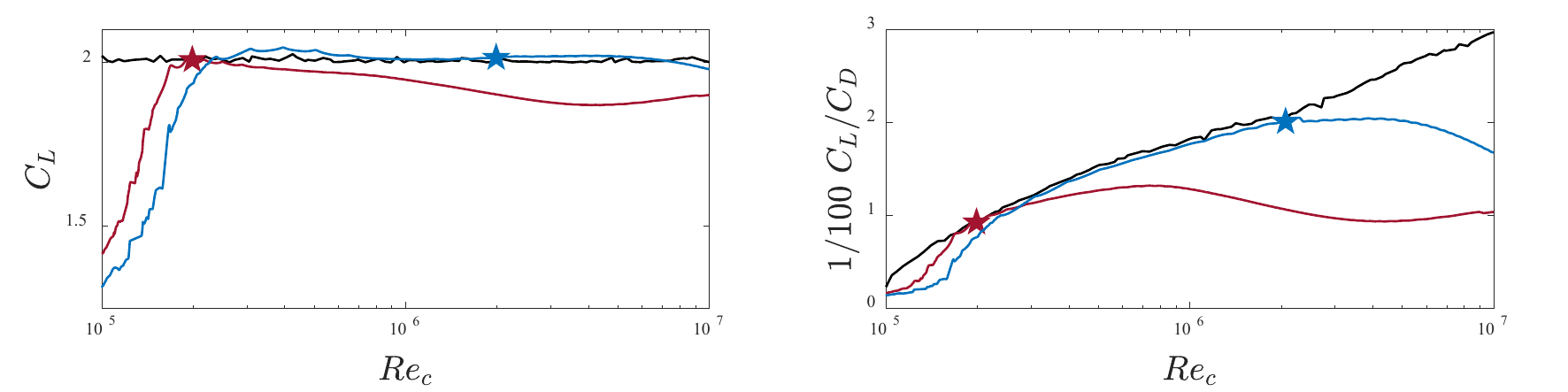}
		\caption{$C_{L}$ and $C_{L}/C_{D}$ of optimal shapes according to $Re_{c}$. The black lines correspond to the optimal shapes obtained in (a). The red and blue lines correspond to each optimal airfoil at $Re_{c}=2\times10^{5}$ and $Re_{c}=2\times10^{6}$ respectively, depicted as the stars.} \label{fig_airfoil_3d_analysis_b}
	\end{subfigure}
	\begin{subfigure}[t]{\linewidth}
		\includegraphics[width=\textwidth]{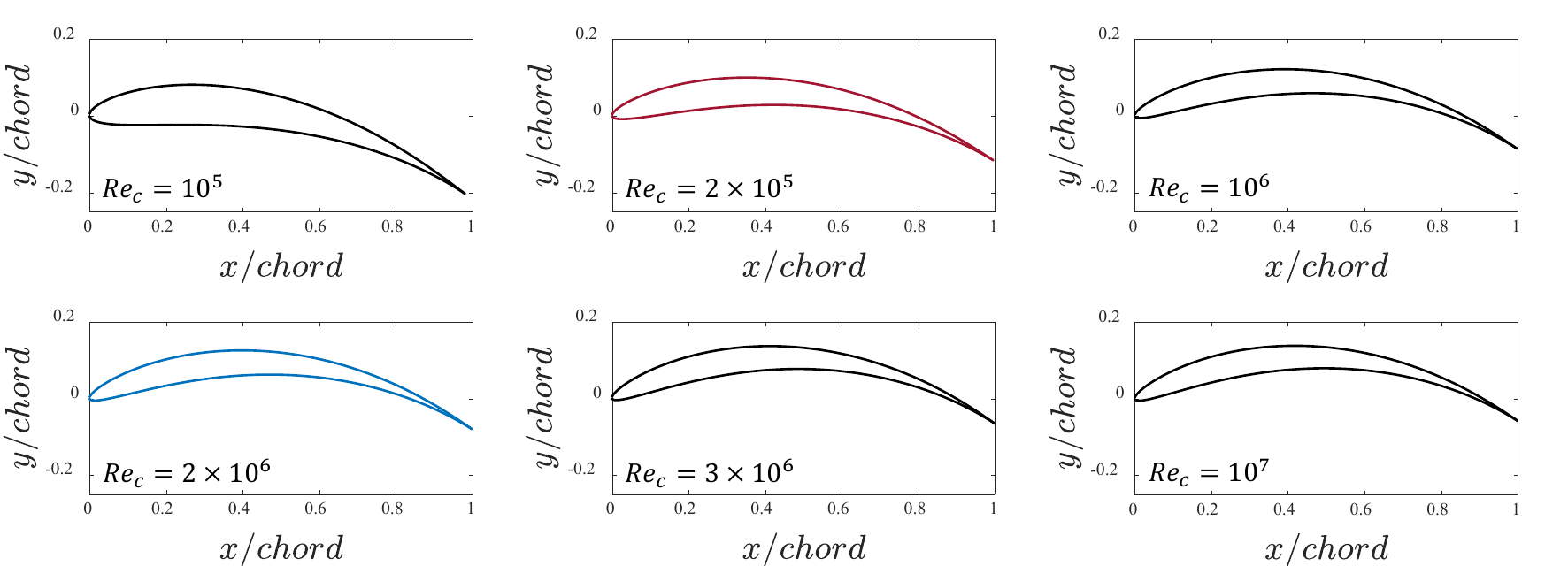}
		\caption{Optimal airfoil shapes obtained in (a) at various $Re_{c}$. Note that the red and blue airfoils correspond to the airfoils selected in (b).} \label{fig_airfoil_3d_analysis_c}
	\end{subfigure}
	\caption{Performance analysis of optimal airfoil shapes.} \label{fig_airfoil_3d_analysis}
	\end{figure}

	In this section, based on the previous optimization results, the need for MC optimization which can be performed over a whole condition space is confirmed. In order to show the need, an analysis is conducted on whether the optimal shapes at some representative conditions can provide sufficient performance over the entire condition space. For the analysis, the optimal airfoil shapes that maximize $C_{L}/C_{D}$ with a constraint $C_{L} \geq 2$ are selected. There are many situations to optimize one objective and keep others above a certain level. For example, in many aviation applications, $C_{L}/C_{D}$ is optimized while maintaining a certain level of $C_{L}$ to sustain their weight\cite{huyse2002probabilistic, RN379, RN380}. The optimal solutions of the constrained optimization problem can be easily obtained from Pareto front as shown in Fig.~\ref{fig_airfoil_3d_analysis_a}.
	
	As shown in the black lines in Fig.~\ref{fig_airfoil_3d_analysis_b}, the optimal shapes show $C_{L}$ greater than $2$ for all $Re_{c}$ and the maximized $C_{L}/C_{D}$ increasing with $Re_{c}$. Then, two optimal shapes at different conditions are used for the analysis. The red lines show $C_{L}$ and $C_{L}/C_{D}$ of the optimal airfoil at $Re_{c}=2\times10^{5}$. Compared to the optimal performance, both $C_{L}$ and $C_{L}/C_{D}$ decrease notably except for the optimized condition. In particular, $C_{L}$ drops significantly at $Re_{c}$ slightly lower than the optimized condition, so the constraint cannot be satisfied at all. In the same way, the blue lines show $C_{L}$ and $C_{L}/C_{D}$ of the optimal airfoil at $Re_{c}=2\times10^{6}$. Although it satisfies the constraint near the optimized condition, it also shows a substantial decrease in $C_{L}/C_{D}$ at $Re_{c}$ slightly higher than the optimized condition.
	
	Fig.~\ref{fig_airfoil_3d_analysis_c} shows the optimal airfoil shapes according to various conditions. The optimal shape at $Re_{c}=2\times10^{5}$ quite differs from the optimal shape at $Re_{c}=10^{5}$, which results in the aforementioned difference in $C_{L}$. However, The optimal shape at $Re_{c}=2\times10^{6}$ is very similar to the optimal shape at $Re_{c}=3\times10^{6}$, although there is a large difference in $C_{L}/C_{D}$ as mentioned before. Likewise, except for the optimal shape at $Re_{c}=10^{5}$, there is no noticeable difference among the other optimal shapes while there exist significant performance differences. These results are attributed to the nonlinear characteristic of a fluid. The optimal shape can drastically change according to the condition, and even if there is no noticeable difference in shape, a slight variation in shape can make a huge performance difference.
	
	Through the analysis, it is shown that an optimal shape at a specific condition cannot be valid at the nearby conditions, and it can be more severe in problems that have nonlinear characteristics. Therefore, it is inadequate to perform optimization by discretizing the condition space into several representative conditions. In order to overcome the problem and remain optimal for varying conditions, it is essential to consider the whole condition space through the MC optimization method proposed in the present study.
	
	\section{Concluding remarks} \label{sec6}
	For the first time in the literature, a MCMO optimization method based on DRL has been developed to find Pareto front over a prescribed condition space. The main idea is based on that DRL can learn a policy for finding optimal solutions according to varying conditions and objectives. The method has been applied to two MCMO optimization problems. First, as a benchmark problem, the Kursawe test function has been newly modified to a MCMO optimization problem. Second, an airfoil shape optimization problem has been dealt with as a practical engineering application. The present MCMO optimization method shows outstanding ability in finding high-resolution Pareto front within the entire condition space including nonlinear and nonconvex parts.
	
	Two additional analyses have been conducted to show its exclusive capability. Firstly, a computational experiment based on the modified Kursawe test function has been carried out to show the effectiveness of MC optimization. Compared with multiple operations of SC optimization for multiple conditions, the number of function evaluations required to find Pareto front is significantly reduced. This efficient optimization is enabled by learning the correlations between conditions and optimal solutions. Secondly, the necessity for MC optimization has been confirmed through an analysis of aerodynamic performance of airfoils with optimally designed shapes. An optimal solution at a specific condition cannot be valid at the nearby conditions, resulting in significant deterioration of target performance. Thus, it is essential to cover the entire condition space, which is possible through the proposed MC optimization method.
	
	The proposed method can show its outstanding capability in optimization problems where conditions and objectives are not fixed. A representative example is shape optimization involving fluid mechanics in which the operating conditions are generally given as a range and the objectives differ depending on the situations. However, the proposed method is not limited to shape optimization and the dimensions of conditions and objectives are not restricted. It can be applied to any MCMO optimization problems. Through the developed MCMO optimization method, it is expected that the fields to which optimization can be practically applied will be greatly expanded. Moreover, from a methodological point of view, this study will pave the way to a new category of optimization as the first MCMO optimization method.
    
    \section*{Declaration of competing interest}
	The authors declare that they have no known competing financial interests or personal relationships that could have appeared to influence the work reported in this paper.
    
	\section*{Acknowledgements}
	The work was supported by the National Research Foundation of Korea (NRF) under the Grant Number NRF-2021R1A2C2092146 and the Samsung Research Funding Center of Samsung Electronics under Project Number SRFC-TB1703-51.
	
	\newpage
    \bibliographystyle{elsarticle-num}
    \biboptions{sort&compress}
    \bibliography{MCMOO}

\begin{thebibliography}{10}
\expandafter\ifx\csname url\endcsname\relax
  \def\url#1{\texttt{#1}}\fi
\expandafter\ifx\csname urlprefix\endcsname\relax\def\urlprefix{URL }\fi
\expandafter\ifx\csname href\endcsname\relax
  \def\href#1#2{#2} \def\path#1{#1}\fi

\bibitem{chong2004introduction}
E.~K. Chong, S.~H. Zak, An introduction to optimization, John Wiley \& Sons,
  2004.

\bibitem{RN355}
J.~Semmler, L.~Pflug, M.~Stingl, G.~Leugering, Shape optimization in
  electromagnetic applications, Springer International Publishing, 2015, pp.
  251--269.

\bibitem{RN408}
S.~Chu, M.~Xiao, L.~Gao, Y.~Zhang, J.~Zhang, Robust topology optimization for
  fiber-reinforced composite structures under loading uncertainty, Computer
  Methods in Applied Mechanics and Engineering 384 (2021) 113935.

\bibitem{RN387}
D.~Taylor, J.-H. Dirks, Shape optimization in exoskeletons and endoskeletons: a
  biomechanics analysis, Journal of The Royal Society Interface 9~(77) (2012)
  3480--3489.

\bibitem{RN399}
J.~Park, A.~Sutradhar, J.~J. Shah, G.~H. Paulino, Design of complex bone
  internal structure using topology optimization with perimeter control,
  Computers in Biology and Medicine 94 (2018) 74--84.

\bibitem{RN386}
B.~Mohammadi, O.~Pironneau, Shape optimization in fluid mechanics, Annual
  Review of Fluid Mechanics 36~(1) (2004) 255--279.

\bibitem{RN390}
G.~Droandi, G.~Gibertini, Aerodynamic blade design with multi-objective
  optimization for a tiltrotor aircraft, Aircraft Engineering and Aerospace
  Technology: An International Journal 87~(1) (2015) 19--29.

\bibitem{RN388}
D.~Peri, M.~Rossetti, E.~F. Campana, Design optimization of ship hulls via
  {CFD} techniques, Journal of Ship Research 45~(02) (2001) 140--149.

\bibitem{RN409}
S.~Percival, D.~Hendrix, F.~Noblesse, Hydrodynamic optimization of ship hull
  forms, Applied Ocean Research 23~(6) (2001) 337--355.

\bibitem{RN381}
S.-H. Yun, Y.-C. Ku, J.-H. Rho, D.-H. Lee, Application of function based design
  method to automobile aerodynamic shape optimization, Multidisciplinary
  Analysis Optimization Conferences, American Institute of Aeronautics and
  Astronautics, 2008.

\bibitem{RN396}
W.~Xudong, W.~Z. Shen, W.~J. Zhu, J.~N. Sørensen, C.~Jin, Shape optimization
  of wind turbine blades, Wind Energy 12~(8) (2009) 781--803.

\bibitem{RN402}
D.~Bertetta, S.~Brizzolara, S.~Gaggero, M.~Viviani, L.~Savio, {CPP} propeller
  cavitation and noise optimization at different pitches with panel code and
  validation by cavitation tunnel measurements, Ocean Engineering 53 (2012)
  177--195.

\bibitem{RN397}
X.~Lachenal, S.~Daynes, P.~M. Weaver, Review of morphing concepts and materials
  for wind turbine blade applications, Wind Energy 16~(2) (2013) 283--307.

\bibitem{RN382}
M.~Secanell, A.~Suleman, P.~Gamboa, Design of a morphing airfoil using
  aerodynamic shape optimization, AIAA Journal 44~(7) (2006) 1550--1562.

\bibitem{RN356}
S.~Vasista, O.~Mierheim, M.~Kintscher, Morphing structures, applications of,
  Springer Berlin Heidelberg, 2019, pp. 1--13.

\bibitem{RN410}
C.~G. Diaconu, P.~M. Weaver, F.~Mattioni, Concepts for morphing airfoil
  sections using bi-stable laminated composite structures, Thin-Walled
  Structures 46~(6) (2008) 689--701.

\bibitem{RN413}
S.~Barbarino, O.~Bilgen, R.~M. Ajaj, M.~I. Friswell, D.~J. Inman, A review of
  morphing aircraft, Journal of Intelligent Material Systems and Structures
  22~(9) (2011) 823--877.

\bibitem{RN411}
R.~M. Ajaj, C.~S. Beaverstock, M.~I. Friswell, Morphing aircraft: The need for
  a new design philosophy, Aerospace Science and Technology 49 (2016) 154--166.

\bibitem{RN389}
N.~Garg, G.~K.~W. Kenway, Z.~Lyu, J.~R. R.~A. Martins, Y.~L. Young,
  High-fidelity hydrodynamic shape optimization of a {3-D} hydrofoil, Journal
  of Ship Research 59~(04) (2015) 209--226.

\bibitem{RN403}
M.~Sacher, M.~Durand, {\'E}.~Berrini, F.~Hauville, R.~Duvigneau, O.~Le~Maître,
  J.-A. Astolfi, Flexible hydrofoil optimization for the 35th {America's Cup}
  with constrained {EGO} method, Ocean Engineering 157 (2018) 62--72.

\bibitem{RN401}
F.~Chen, L.~Liu, X.~Lan, Q.~Li, J.~Leng, Y.~Liu, The study on the morphing
  composite propeller for marine vehicle. part {I}: {Design} and numerical
  analysis, Composite Structures 168 (2017) 746--757.

\bibitem{RN391}
N.~Srinivas, K.~Deb, Muiltiobjective optimization using nondominated sorting in
  genetic algorithms, Evolutionary Computation 2~(3) (1994) 221--248.

\bibitem{RN392}
K.~Deb, A.~Pratap, S.~Agarwal, T.~Meyarivan, A fast and elitist multiobjective
  genetic algorithm: {NSGA-II}, IEEE Transactions on Evolutionary Computation
  6~(2) (2002) 182--197.

\bibitem{RN393}
C.~Coello~Coello, M.~Lechuga, {MOPSO}: {A} proposal for multiple objective
  particle swarm optimization, in: Proceedings of the 2002 Congress on
  Evolutionary Computation. CEC'02 (Cat. No.02TH8600), Vol.~2, 2002, pp.
  1051--1056.

\bibitem{RN357}
K.~Miettinen, M.~M. Mäkelä, On scalarizing functions in multiobjective
  optimization, OR Spectrum 24~(2) (2002) 193--213.

\bibitem{RN398}
W.~Wang, Y.~Li, M.~K. Osman, S.~Yuan, B.~Zhang, J.~Liu, Multi-condition
  optimization of cavitation performance on a double-suction centrifugal pump
  based on {ANN} and {NSGA-II}, Processes 8~(9) (2020) 1124.

\bibitem{RN412}
X.~Yan, J.~Zhu, M.~Kuang, X.~Wang, Aerodynamic shape optimization using a novel
  optimizer based on machine learning techniques, Aerospace Science and
  Technology 86 (2019) 826--835.

\bibitem{RN383}
J.~Li, M.~Zhang, J.~R. R.~A. Martins, C.~Shu, Efficient aerodynamic shape
  optimization with deep-learning-based geometric filtering, AIAA Journal
  58~(10) (2020) 4243--4259.

\bibitem{RN358}
J.~Rabault, F.~Ren, W.~Zhang, H.~Tang, H.~Xu, Deep reinforcement learning in
  fluid mechanics: A promising method for both active flow control and shape
  optimization, Journal of Hydrodynamics 32~(2) (2020) 234--246.

\bibitem{RN400}
J.~Viquerat, J.~Rabault, A.~Kuhnle, H.~Ghraieb, A.~Larcher, E.~Hachem, Direct
  shape optimization through deep reinforcement learning, Journal of
  Computational Physics 428 (2021) 110080.

\bibitem{RN385}
S.~Qin, S.~Wang, L.~Wang, C.~Wang, G.~Sun, Y.~Zhong, Multi-objective
  optimization of cascade blade profile based on reinforcement learning,
  Applied Sciences 11~(1) (2021) 106.

\bibitem{RN384}
R.~Li, Y.~Zhang, H.~Chen, Learning the aerodynamic design of supercritical
  airfoils through deep reinforcement learning, AIAA Journal 59~(10) (2021)
  3988--4001.

\bibitem{Kursawe1990}
F.~Kursawe, A variant of evolution strategies for vector optimization, in:
  Parallel Problem Solving from Nature, Springer Berlin Heidelberg, 1991, pp.
  193--197.

\bibitem{RN407}
X.~Zhang, F.~Xie, T.~Ji, Z.~Zhu, Y.~Zheng, Multi-fidelity deep neural network
  surrogate model for aerodynamic shape optimization, Computer Methods in
  Applied Mechanics and Engineering 373 (2021) 113485.

\bibitem{RN406}
K.~Wang, S.~Yu, Z.~Wang, R.~Feng, T.~Liu, Adjoint-based airfoil optimization
  with adaptive isogeometric discontinuous {Galerkin} method, Computer Methods
  in Applied Mechanics and Engineering 344 (2019) 602--625.

\bibitem{RN405}
E.~Gillebaart, R.~De~Breuker, Low-fidelity 2{D} isogeometric aeroelastic
  analysis and optimization method with application to a morphing airfoil,
  Computer Methods in Applied Mechanics and Engineering 305 (2016) 512--536.

\bibitem{miettinen2012nonlinear}
K.~Miettinen, Nonlinear multiobjective optimization, Vol.~12, Springer Science
  \& Business Media, 2012.

\bibitem{RN394}
Q.~Zhang, H.~Li, {MOEA/D}: A multiobjective evolutionary algorithm based on
  decomposition, IEEE Transactions on Evolutionary Computation 11~(6) (2007)
  712--731.

\bibitem{RN404}
Y.-y. Tan, Y.-c. Jiao, H.~Li, X.-k. Wang, {MOEA/D}+ uniform design: A new
  version of {MOEA/D} for optimization problems with many objectives, Computers
  \& Operations Research 40~(6) (2013) 1648--1660.

\bibitem{RN395}
K.~Van~Moffaert, M.~M. Drugan, A.~Nowé, Scalarized multi-objective
  reinforcement learning: {N}ovel design techniques, in: 2013 IEEE Symposium on
  Adaptive Dynamic Programming and Reinforcement Learning (ADPRL), pp.
  191--199.

\bibitem{sutton2018reinforcement}
R.~S. Sutton, A.~G. Barto, Reinforcement learning: {A}n introduction, MIT
  Press, 2018.

\bibitem{RN359}
R.~Bellman, Dynamic programming, Science 153~(3731) (1966) 34--37.

\bibitem{RN360}
V.~Mnih, K.~Kavukcuoglu, D.~Silver, A.~A. Rusu, J.~Veness, M.~G. Bellemare,
  A.~Graves, M.~Riedmiller, A.~K. Fidjeland, G.~Ostrovski, S.~Petersen,
  C.~Beattie, A.~Sadik, I.~Antonoglou, H.~King, D.~Kumaran, D.~Wierstra,
  S.~Legg, D.~Hassabis, Human-level control through deep reinforcement
  learning, Nature 518~(7540) (2015) 529--533.

\bibitem{mnih2013playing}
V.~Mnih, K.~Kavukcuoglu, D.~Silver, A.~Graves, I.~Antonoglou, D.~Wierstra,
  M.~Riedmiller, Playing atari with deep reinforcement learning (2013).
\newblock \href {http://arxiv.org/abs/1312.5602} {\path{arXiv:1312.5602}}.

\bibitem{RN361}
L.~Buşoniu, T.~de~Bruin, D.~Tolić, J.~Kober, I.~Palunko, Reinforcement
  learning for control: Performance, stability, and deep approximators, Annual
  Reviews in Control 46 (2018) 8--28.

\bibitem{RN362}
P.~Garnier, J.~Viquerat, J.~Rabault, A.~Larcher, A.~Kuhnle, E.~Hachem, A review
  on deep reinforcement learning for fluid mechanics, Computers \& Fluids 225
  (2021) 104973.

\bibitem{viquerat2021review}
J.~Viquerat, P.~Meliga, E.~Hachem, A review on deep reinforcement learning for
  fluid mechanics: an update (2021).
\newblock \href {http://arxiv.org/abs/2107.12206} {\path{arXiv:2107.12206}}.

\bibitem{konda2000actor}
V.~R. Konda, J.~N. Tsitsiklis, Actor-critic algorithms, in: Advances in Neural
  Information Processing Systems, 2000, pp. 1008--1014.

\bibitem{maas2013rectifier}
A.~L. Maas, A.~Y. Hannun, A.~Y. Ng, Rectifier nonlinearities improve neural
  network acoustic models, in: Proceedings of the 30th International Conference
  on Machine Learning, Vol.~30, Citeseer, 2013, p.~3.

\bibitem{kingma2017adam}
D.~P. Kingma, J.~Ba, Adam: {A} method for stochastic optimization (2017).
\newblock \href {http://arxiv.org/abs/1412.6980} {\path{arXiv:1412.6980}}.

\bibitem{RN363}
S.~Fujimoto, H.~van Hoof, D.~Meger, Addressing function approximation error in
  actor-critic methods, in: Proceedings of the 35th International Conference on
  Machine Learning, Vol.~80, PMLR, 2018, pp. 1587--1596.

\bibitem{zitzler1999evolutionary}
E.~Zitzler, Evolutionary algorithms for multiobjective optimization: {M}ethods
  and applications, Vol.~63, Citeseer, 1999.

\bibitem{RN364}
A.~Auger, J.~Bader, D.~Brockhoff, E.~Zitzler, Hypervolume-based multiobjective
  optimization: {T}heoretical foundations and practical implications,
  Theoretical Computer Science 425 (2012) 75--103.

\bibitem{RN365}
W.~J. Lim, A.~B. Jambek, S.~C. Neoh, Kursawe and {ZDT} functions optimization
  using hybrid micro genetic algorithm {(HMGA)}, Soft Computing 19~(12) (2015)
  3571--3580.

\bibitem{RN367}
C.~J. Tan, C.~P. Lim, Y.-N. Cheah, A modified micro genetic algorithm for
  undertaking multi-objective optimization problems, Journal of Intelligent \&
  Fuzzy Systems 24 (2013) 483--495.

\bibitem{RN368}
M.-F. Leung, S.-C. Ng, C.-C. Cheung, A.~K. Lui, A new strategy for finding good
  local guides in {MOPSO}, in: 2014 IEEE Congress on Evolutionary Computation
  (CEC), 2014, pp. 1990--1997.

\bibitem{RN369}
Y.~Naranjani, C.~Hernández, F.-R. Xiong, O.~Schütze, J.-Q. Sun, A hybrid
  method of evolutionary algorithm and simple cell mapping for multi-objective
  optimization problems, International Journal of Dynamics and Control 5~(3)
  (2017) 570--582.

\bibitem{RN370}
R.~Mukesh, K.~Lingadurai, U.~Selvakumar, Airfoil shape optimization using
  non-traditional optimization technique and its validation, Journal of King
  Saud University - Engineering Sciences 26~(2) (2014) 191--197.

\bibitem{RN371}
A.~F.~P. Ribeiro, A.~M. Awruch, H.~M. Gomes, An airfoil optimization technique
  for wind turbines, Applied Mathematical Modelling 36~(10) (2012) 4898--4907.

\bibitem{RN372}
S.~Zhang, H.~Li, W.~Jia, D.~Xi, Multi-objective optimization design for
  airfoils with high lift-to-drag ratio based on geometric feature control, IOP
  Conference Series: Earth and Environmental Science 227 (2019) 032014.

\bibitem{huyse2002probabilistic}
L.~Huyse, S.~L. Padula, R.~M. Lewis, W.~Li, Probabilistic approach to free-form
  airfoil shape optimization under uncertainty, AIAA journal 40~(9) (2002)
  1764--1772.

\bibitem{milne1973theoretical}
L.~M. Milne-Thomson, Theoretical aerodynamics, Courier Corporation, 1973.

\bibitem{Paola2012}
P.~Puorger, D.~Dessi, F.~Mastroddi, Preliminary design of an amphibious
  aircraft by the multidisciplinary design optimization approach, American
  Institute of Aeronautics and Astronautics, 2007, p. 1924.

\bibitem{RN373}
M.~Berci, V.~V. Toropov, R.~W. Hewson, P.~H. Gaskell, Multidisciplinary
  multifidelity optimisation of a flexible wing aerofoil with reference to a
  small {UAV}, Structural Multidisciplinary Optimization 50~(4) (2014)
  683–699.

\bibitem{RN374}
M.~Drela, {XFOIL}: {A}n analysis and design system for low {R}eynolds number
  airfoils, in: Low Reynolds Number Aerodynamics, Springer Berlin Heidelberg,
  1989, pp. 1--12.

\bibitem{RN375}
T.~H. Hansen, Airfoil optimization for wind turbine application, Wind Energy
  21~(7) (2018) 502--514.

\bibitem{RN376}
S.~Zhang, H.~Li, A.~A. Abbasi, Design methodology using characteristic
  parameters control for low {R}eynolds number airfoils, Aerospace Science and
  Technology 86 (2019) 143--152.

\bibitem{RN377}
K.~R. Ram, S.~P. Lal, M.~R. Ahmed, Design and optimization of airfoils and a 20
  k{W} wind turbine using multi-objective genetic algorithm and {HARP\_Opt}
  code, Renewable Energy 144 (2019) 56--67.

\bibitem{RN378}
P.~B.~S. Lissaman, Low-{R}eynolds-number airfoils, Annual Review of Fluid
  Mechanics 15~(1) (1983) 223--239.

\bibitem{RN379}
H.~P. Buckley, B.~Y. Zhou, D.~W. Zingg, Airfoil optimization using practical
  aerodynamic design requirements, Journal of Aircraft 47~(5) (2010)
  1707--1719.

\bibitem{RN380}
M.~Nemec, D.~W. Zingg, T.~H. Pulliam, Multipoint and multi-objective
  aerodynamic shape optimization, AIAA Journal 42~(6) (2004) 1057--1065.

\end{thebibliography}
	
	\newpage
    \begin{algorithm}[H]
    {\footnotesize
        \label{alg_drl}
	    \caption{Single-step DRL for MCMO optimization}
		initialize actor network $\pi_{\phi}$ and critic network $Q_{\theta}$ with random parameters $\phi$, $\theta$;\\
		initialize episode, utopia $\bf{f}^*$, and replay buffer $\mathcal{B}$;\\
		\Repeat{convergence}{
		    episode $\gets$ episode+1;\\
    		select condition and weight: $\bf{c}\sim \text{U}(0,1)$, $\bf{w}\sim \text{U}(0,1)$ $s.t.$ $|\bf{w}|=1$;\\
    		calculate utopia $\bf{f}^*(\bf{c})$;\\
    		receive state: $\bf{s}\gets[\bf{c},\bf{w},\bf{f}^*]$;\\
    		select action with exploration noise: $\bf{a}\gets$clip$(\pi_{\phi}(\bf{s})+\bf{\epsilon},-1,1)$, $\bf{\epsilon}\sim\mathcal{N}(0,\sigma^{2})$;\\
    		execute action and evaluate objective function $\bf{f}(\bf{x},\bf{c})$;\\
            $\forall i$, if $f_i(\bf{x},\bf{c})<f_i^*(\bf{c})$, update utopia and state;\\
            calculate reward and reproduce data;\\
            store data of $(\bf{s},\bf{a},r)$ in $\mathcal{B}$;\\

    		\For{$l=1$ \KwTo $l_{max}$}{
        		sample mini-batch of $N_b$ data from $\mathcal{B}$;\\
        		update $\theta$ with the loss $N_b^{-1}\sum(r-Q_{\theta}(\bf{s},\bf{a}))^2$;\\			
        		\If{$l$ mod $l_d$}{
            		update $\phi$ by the deterministic policy gradient $N_b^{-1}\sum\nabla_{\bf{a}} Q_{\theta}(\bf{s},\bf{a})|_{\bf{a}=\pi_{\phi}(\bf{s})}\nabla_{\phi}\pi_{\phi}(\bf{s})$;\\
        		}
        	
        	}
		}
	}
    \end{algorithm}
	
	\newpage
	\listoffigures
	\pagebreak
	\clearpage
	
	\pagebreak
	\clearpage	
	
	\pagebreak
	\clearpage	

	\pagebreak
	\clearpage

    \pagebreak
    \clearpage
	
	\pagebreak
	\clearpage
	
	\pagebreak
	\clearpage
	
	\pagebreak
	\clearpage

\pagebreak
\clearpage

\thispagestyle{empty}
\section*{Highlights}
\begin{sloppypar}
\begin{itemize}

	\item{
		A multi-condition multi-objective optimization method is developed based on deep reinforcement learning.
	}
	\item{
		A novel benchmark problem for multi-condition multi-objective optimization is introduced.
	}
	\item{
		The developed method is shown to efficiently find high-resolution Pareto front over a condition space.
	}
	\item{
		Learning the correlations between conditions and optimal solutions enables efficient optimization.
	}
	\item{
	    Critical degradation of target performance by optimization performed only at a specific condition is confirmed.
	}
	
\end{itemize}
\end{sloppypar}
\pagebreak
\clearpage

\end{document}